\pdfoutput=1

\documentclass[11pt]{article}

\usepackage[final]{acl}

\usepackage{times}
\usepackage{latexsym}
\usepackage{longtable}  
\usepackage[T1]{fontenc}

\usepackage[utf8]{inputenc}

\usepackage{microtype}

\usepackage{inconsolata}

\usepackage{graphicx}

\usepackage{hyperref}
\usepackage{wrapfig}
\usepackage{url}
\usepackage{amssymb}
\usepackage{float}
\usepackage{booktabs}
\usepackage{multirow}
\usepackage{algorithm}
\usepackage{algorithmic}
\usepackage{svg}
\usepackage{colortbl}
\usepackage{wrapfig}
\usepackage{rotating}
\usepackage[ragged]{sidecap}  
\usepackage{amsmath}
\usepackage{cleveref}
\usepackage{placeins}
\DeclareUnicodeCharacter{2212}{\ensuremath{-}}
\usepackage{xcolor}
\definecolor{darkgreen}{rgb}{0.0, 0.65, 0.0}
\definecolor{darkorange}{rgb}{1.0, 0.55, 0.0}
\definecolor{cobalt}{rgb}{0.0, 0.28, 0.67}
\definecolor{high}{rgb}{1,0,0}
\usepackage[most]{tcolorbox}
\newcommand{\intuition}[1]{
\begin{tcolorbox}[colback=white,boxrule=1pt,top=0pt,bottom=0pt,left=1pt,right=2pt,top=2pt,bottom=2pt]
\em #1
\end{tcolorbox}
}

\title{Exploring Model Kinship for Merging Large Language Models}

\author{
    Yedi Hu$^{\spadesuit}$,
    Yunzhi Yao$^{\spadesuit}$,
    Ningyu Zhang$^\spadesuit$,
    Huajun Chen$^{\spadesuit}$, 
    Shumin Deng$^\diamondsuit$\thanks{Corresponding Author.} \\
    $^\spadesuit$Zhejiang University ~
    $^\diamondsuit$National University of Singapore, NUS-NCS Joint Lab \\
    \texttt{\{231sm,zhangningyu\}@zju.edu.cn}
    \quad
    \texttt{shumin@nus.edu.sg}
}

\begin{document}
\maketitle
\begin{abstract}
Model merging has emerged as a key technique for enhancing the capabilities and efficiency of Large Language Models (LLMs). The open-source community has driven model evolution by iteratively merging existing models, yet a principled understanding of the gains and underlying factors in
model merging remains limited. In this work, we study model evolution through iterative merging, drawing an analogy to \textbf{biological evolution}, and introduce the concept of \textit{model kinship}, the degree of similarity or relatedness between LLMs. Through comprehensive empirical analysis, we show that model kinship is closely linked to the performance improvements achieved by merging, providing a useful criterion for selecting candidate models. Building on this insight, we propose a new model merging strategy: Top-$k$ Greedy Merging with Model Kinship, which can improve benchmark performance. Specifically, we discover that incorporating model kinship as a guiding criterion enables continuous merging while mitigating performance degradation caused by local optima, thereby facilitating more effective model evolution\footnote{Code is available at \url{https://github.com/zjunlp/ModelKinship}.}. 
\end{abstract}

\section{Introduction}
Fine-tuning pre-trained models (PTMs) for downstream tasks has become a popular practice, and has proven particularly effective for Large Language Models (LLMs) \citep{Kol2020, qiu2021, askell2021, ouyang2022,zhao2023}. 
However, deploying separate fine-tuned models for each task can be resource-intensive \citep{fifty2021efficiently}, which drives  increasing demands for multi-task learning solutions \citep{Zhang2021Survey, arXiv2024_Survey-Multi-Task-Learning, mec2024, cool2024}. 

\begin{figure}[!ht]
    \centering
    \includegraphics[width=\linewidth]{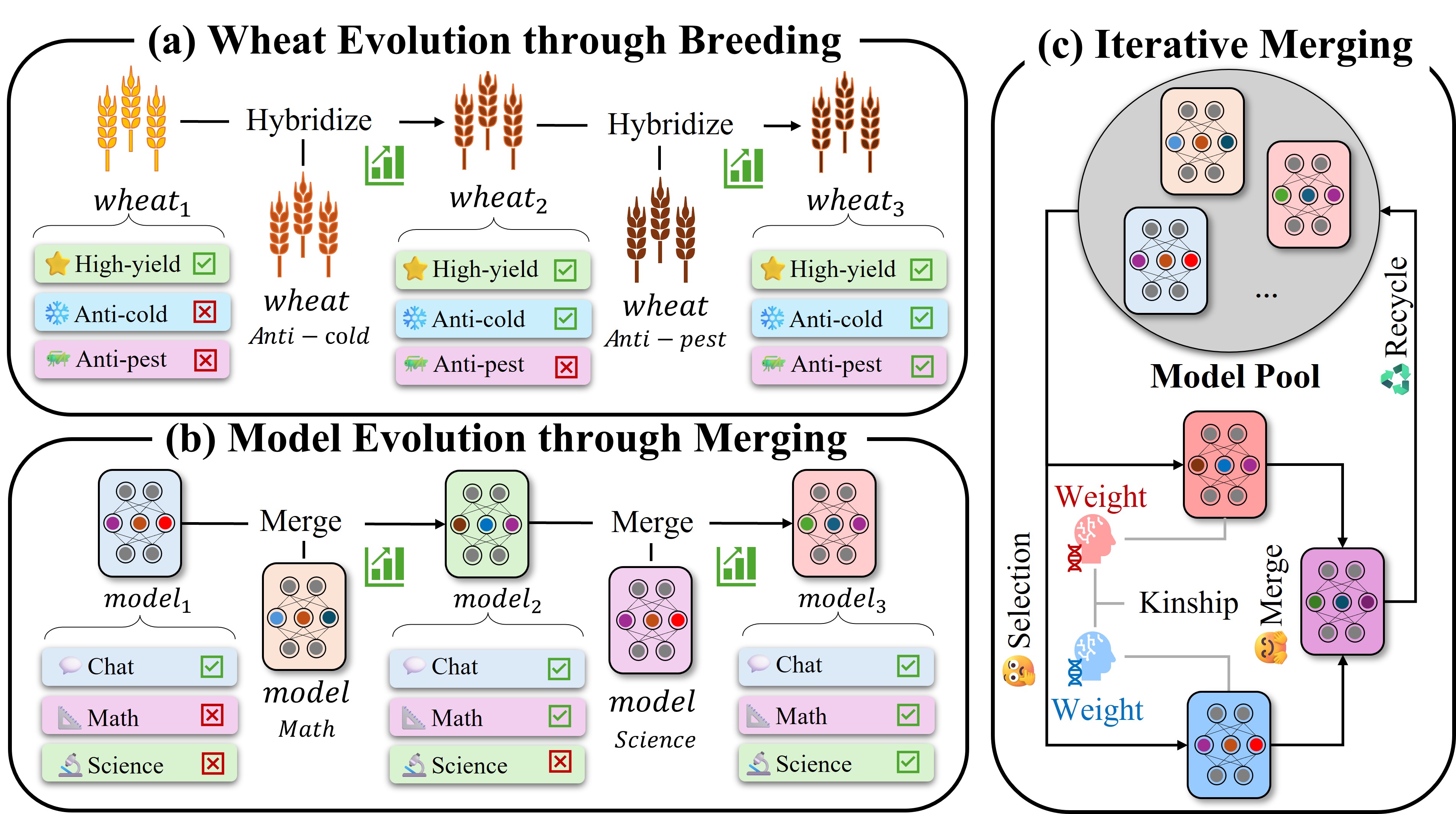}
      \vspace{-13pt}
    \caption{\textbf{An intuitive comparison between wheat evolution and model evolution}. 
    A parallel can be drawn between biological reproduction \textbf{(Part a)} and the process of model evolution \textbf{(Part b)}. 
    In biological systems, offspring inherit genetic material from both parents, forming a new genotype through the combination of parental traits. 
    Similarly, in model merging, the merged model inherits parameters or weights from the contributing models. \textbf{Part c} demonstrates the iterative execution of model evolution.
    Starting with a group of LLMs, the repository evolves through a Selection-Merge-Recycle iteration. 
    Notably, \textit{model kinship} can serve as an effective tool to guide this iterative model merging process (e.g., infer whether there may be gains after model merging.).}
    \vspace{-10pt}
    \label{fig:compare}
\end{figure}

Recent studies suggest that model merging \citep{SinghJ20, sung2023, goddard2024arcee, matena2022, yang2024model, DBLP:conf/eccv/JangYH24} offers a viable approach for achieving multi-task objectives by integrating multiple expert models. 
Furthermore, advances in model merging toolkits \citep{goddard2024arcee, Tang2024} have lowered the technical barrier, allowing users with limited expertise to easily conduct merging experiments, thus leading to an evolution of LLMs for the community. 

At present, researchers have developed various powerful LLMs using model merging techniques \citep{open-llm-leaderboard}. 
Many of these models are created through a biologically inspired evolutionary process  involving iterative merging, an approach that we refer to as model evolution (Figure~\ref{fig:compare}(a,b)).
Despite these successes,  current merging practice faces critical limitations. Progress often relies on continuous trial and extensive human expertise, with little formal guidance or standardized procedure. 
\textbf{To address this problem, we propose an iterative model merging framework (Figure~\ref{fig:compare}(c))}, leveraging explicit strategies to guide the direction of model evolution toward improved performance. We show that even a simple greedy strategy can outperform baseline merging approaches.

However, in the later stages of both community-driven model merging experiments and greedy evolution strategies, achieving additional gains in multi-task capability becomes increasingly challenging.
To explore possible solutions, we introduce `\textbf{model kinship}', a metric inspired by the concept of kinship in evolutionary biology \citep{sahlins2013kinship}, to inform and enhance the merging process.
Model kinship is designed to quantify the degree of similarity or relationship between models throughout the iterative merging process.
By offering a principled framework for measuring these relationships, model kinship provides valuable insights that can refine merging strategies for model evolution.

We conduct a comprehensive analysis of model merging experiments based on model kinship.
We observe that the  model merging process consists of two distinct stages: (1) an improving stage, where models exhibit significant performance gains, and (2) a saturation stage, where improvements diminish and eventually plateau.
Empirically, we find \textbf{a strong correlation between model kinship and variations in average task performance}, suggesting that model kinship is indicative of potential effectiveness for model merging.
These findings inspire two main insights: \textbf{(1) high-kinship merges can lead to performance stagnation, akin to inbreeding; (2) low-kinship merges carry greater risk but may yield larger gains and facilitate escape from local optima.}

Inspired by this, we propose a novel continual model merging strategy: \emph{Top-$k$ Greedy Merging with Model Kinship}. 
Specifically, we find that leveraging model kinship as a criterion enables more effective model merging, helping mitigate degradation and avoid local optima during model evolution. 
Furthermore, model kinship also proves useful as an early stopping criterion, improving efficiency of the merging process.  

In general, this paper mainly contains \textbf{four key contributions}:

\begin{enumerate}

    \item \textbf{Iterative Model Merging as a Framework for Model Evolution:} We propose continual model merging as a viable framework for evolving LLMs. Through strategically guided merging across iterations, this approach yields consistent improvements in generalization and task performance.

    \item \textbf{Introducing \textit{Model Kinship}:} We introduce model kinship, designed to assess the degree of similarity or relationship between LLMs during the merging process, which can guide model merging strategies and holds promise for advancing auto merging research.

    \item \textbf{Empirical Analysis of Model Evolution:} We present a comprehensive empirical analysis of model evolution through iterative merging.  
    Our findings highlight the dynamics of multi-task performance improvement and stagnation during evolution. 
    In addition, we propose a preliminary explanation of the underlying mechanisms using model kinship.

    \item \textbf{Practical Model Merging Strategies with \textit{Model Kinship}:} We demonstrate how model kinship guides the model merging process to tackle optimization challenges, and provide practical strategies: Top-$k$ Greedy Merging with Model Kinship, to enhance efficiency and effectiveness of model evolution.

\end{enumerate}

\section{Background}

\label{sec:background}
\subsection{Model Merging: Fundamentals}

Model merging aims to integrate two or more domain-specific models into a unified framework, thereby harnessing their composite capabilities across multiple tasks \citep{sung2023}.
While this approach shares conceptual similarities with ensemble methods \citep{dietterich2002ensemble,dong2020survey,jiang-etal-2023-llm}, model merging generates a single, generalized model, avoiding the increased inference time associated with ensembles. 
Let \(f_i\) represent the \(i\)-th model for merging, each with its unique parameters \(\theta_i\). If the merging process follows method \(\mathcal{F}\), the prediction \(\hat{y}\) of the merged model \(f_{\text{merge}}\) for input \(x\) is:

\vspace{-10pt}

{\small
\begin{align}
\hat{y} = f_{\text{merge}}(x) = \mathcal{F}\left( f_1(x; \theta_1), f_2(x; \theta_2), \ldots, f_n(x; \theta_n) \right)
\end{align}
}

\subsection{Model Evolution: Benefits and Challenges}
Parameter averaging methods allow merged models to retain the same architecture and parameter size as their original components, enabling reuse in subsequent merging steps.
Leveraging this property, the open-source community has progressively enhanced model performance through repeated merging, leading to a ``\textbf{Model Evolution}''.
Empirical evidence from the Open LLM Leaderboard \citep{open-llm-leaderboard} shows that this iterative model merging process can produce highly generalized models, often outperforming those produced through a single merging step \citep{yamshadow_2024}.

Despite these advances, current community practices largely rely on random merging by multiple contributors, which results in high computational costs and unstable behavior, limiting their practicality for industrial applications.

\section{Iterative Model Merging and Strategy Boost}
\label{sec:model_evolution}

In this section, we present controlled experiments to demonstrate that \textbf{iterative model merging}, the process of repeatedly combining models, can help stabilize improvements in multi-task capability. Moreover, we demonstrate that incorporating a \textbf{selection strategy}, such as Top-$k$ Greedy, yields more substantial performance improvements.

\subsection{Iterative Model Merging Framework}

We formally define \textbf{iterative model merging} as a modular framework that evolves a population of models through repeated merging. It is governed by three components: a \textbf{selection strategy} \( \mathcal{S} \), which selects candidate models from the pool \( \mathcal{P}_t \); a \textbf{merging operator} \( \mathcal{M} \), such as SLERP or weighted averaging, to combine the selected models; and an \textbf{stopping criterion} \( \mathcal{E} \), which determines when to stop the process.

At each generation \( t \), a subset \( \mathcal{S}_t \subseteq \mathcal{P}_t \) is selected using \( \mathcal{S} \), merged via \( \mathcal{M} \), and the resulting model \( M_{t+1} \) is added back to the pool. The process continues until the stopping criterion \( \mathcal{E} \) is met:
\begin{equation}
M_{t+1} = \mathcal{M}(\mathcal{S}(\mathcal{P}_t)) \quad \text{until} \quad \mathcal{E}(M_{t+1})
\end{equation}

\begin{table*}[t]
  \captionsetup{aboveskip=2pt, belowskip=2pt}
  \centering
  \small
  \setlength{\tabcolsep}{3pt} 
  \renewcommand{\arraystretch}{0.9} 
  \begin{tabular*}{0.7\textwidth}{@{\hspace{6pt}\extracolsep{\fill}} l|ccc|c@{\hspace{6pt}}}
    \toprule
    \textbf{Method} & \textbf{TruthfulQA} & \textbf{Winogrande} & \textbf{GSM8k} & \textbf{Avg.} \\
    \midrule
    Ties (Density=0.5, Weight=0.3) & 62.76 & 79.56 & 8.79 & 50.37 \\
    Dare Ties (Density=0.5, Weight=0.3) & 59.36 & 79.08 & 65.73 & 68.05 \\
    Linear (Weight=0.3) & 56.37 & 78.08 & 68.54 & 67.66 \\
    \midrule
    Sequential SLERP Merge (ord1) & 47.15 & 76.24 &  53.15 & 58.84 \\
    Sequential SLERP Merge (ord2) & 61.01 & 79.56 & 63.76 & 68.11 \\
    Sequential SLERP Merge (ord3) & 49.80 & 78.53 & 55.72 & 61.35 \\
    \midrule
    \cellcolor{gray!20} \textbf{Random Merging (k=3)} & \cellcolor{gray!20} 54.32& \cellcolor{gray!20} 78.53 & \cellcolor{gray!20} 72.81 & \cellcolor{gray!20} \textbf{68.55} \\
    \cellcolor{gray!20} \textbf{Top k Greedy Merging (k=3)} &\cellcolor{gray!20}  50.94 & \cellcolor{gray!20} 80.11 & \cellcolor{gray!20} 75.13 & \cellcolor{gray!20} \textbf{68.72} \\
    \bottomrule
  \end{tabular*}
    \caption{\textbf{Performance Comparison} across multi-model merging, sequential model merging with different SLERP orders, and iterative model merging with strategy.}
  \vspace{-4mm}
 \label{tab:merge_comparison}
\end{table*}

\subsection{Setup}

\paragraph{Baseline Methods.}
We consider two types of baselines: (1) \textbf{Multi-model merging} methods that support merging all models at once, such as \textbf{TIES}~\citep{yadav2023}, \textbf{Dare-TIES}~\citep{yu2024}, and \textbf{Linear}; and (2) \textbf{Sequential merging}, where models are merged pairwise using \textbf{SLERP}~\citep{Shoemake1985}.

\paragraph{Top-$k$ Greedy Merging.}
Our approach applies \textbf{iterative model merging} using a \textbf{Top-$k$ greedy} selection strategy on $n$ LLMs (see Algorithm~\ref{alg:greedy_merge}). Each merge step uses SLERP. We also include a \textbf{random merge} baseline (Appendix~\ref{sec:appendix-random}).

\paragraph{Models and Datasets.}
We use three HuggingFace LLMs based on \textit{Mistral-7B}: \textbf{\textit{mistral-7b-instruct-v0.2}}, \textbf{\textit{metamath-mistral-7b}}, and \textbf{\textit{open-chat-3.5-1210}}.

\paragraph{Evaluation.}
Models are evaluated on Winogrande, GSM8k, and TruthfulQA, which highlight their task-specific strengths. Dataset details are in Appendix~\ref{sec:dataset}.

\subsection{Results}
As shown in Table~\ref{tab:merge_comparison}, iterative model merging can yield better generalization when combining multiple tasks. \textbf{In particular, both random and Top-$k$ greedy iterative merging outperform one-step baselines, demonstrating the effectiveness of continual merging.} While single-step merging methods can sometimes achieve strong results (e.g., Linear or Dare Ties), their performance is often unstable and sensitive to hyperparameter. Ties, for instance, fails drastically on GSM8k. Sequential SLERP merging shows similar limitations, as its performance varies significantly depending on the merge order. \textbf{In contrast, iterative merging strategies are more stable, consistently yielding robust performance across tasks.} 

However, these methods offer only marginal improvements. This raises a key question: \textbf{\textit{How can we further enhance performance beyond the current limits of naive or greedy strategies?}}

\section{Preliminary Analysis of Model Kinship}
\label{sec:emprical_results}

To address this limitation, we investigate whether structural signals beyond raw performance can guide model selection. We introduce \textbf{model kinship}, a metric capturing parameter-space similarity, and conduct a preliminary analysis to examine its correlation with merge gains in community-merged LLMs. This helps assess the potential of kinship-aware strategies in guiding continual merging and avoiding local optima.

\subsection{Model Kinship}
Drawing inspiration from the parallel between artificial selection and model evolution (as detailed in Appendix~\ref{biology}), we hypothesize that a concept analogous to \emph{kinship}, the genetic relatedness studied in evolutionary biology \citep{thompson1985}, can also apply to model merging. Specifically, we introduce the notion of \emph{model kinship}, a metric designed to capture and quantify the evolutionary relationships between the merge candidates. This analogy suggests that, as genetic kinship affects breeding outcomes, model kinship similarly influences the effectiveness of merging strategies in enhancing multi-task performance.

We adopt the most intuitive representation, inspired by the cosine similarity analysis introduced in the Task Arithmetic paper \citep{ilharco2023}.
This metric is designed to evaluate the degree of similarity or relatedness between the task capabilities of large language models (LLMs) based solely on their "genetic" information, meaning the changes in their weights, during model evolution.
Considering two models $m_{i}$, $m_{j}$ involved in a model evolution originated from the pre-trained model $m_{base}$, the weights of $m_{i}$, $m_{j}$ are denoted as $\theta_{\text{i}},\theta_{\text{j}} \in \mathbb{R}^d$. 
Similarly, $\theta_{\text{base}} \in \mathbb{R}^d$ represents the weights of the pre-trained model. 
Since the differences between models emerge after fine-tuning and merging, the variation of weights during model evolution is crucial. It is calculated as:

\vspace{-10pt}

\begin{align}
\delta_{\text{i}} = \theta_{\text{i}} - \theta_{\text{base}},  \delta_{\text{j}} = \theta_{\text{j}} - \theta_{\text{base}}
\end{align}

Model kinship $r$ is designed to capture the similarity of task capabilities between models. 
In this paper, we explore multiple potential metrics for evaluating similarity.
For the calculation, $sim(\cdot, \cdot)$ denotes the similarity metric function used. 
Considering two cases merging of 2 models and merging of $n$ models, we define model kinship $r$ as:

\vspace{-10pt}

\begin{align}
r = 
\begin{cases} 
    sim(\delta_1, \delta_2), & \text{(Merge 2)} \\
    \frac{2}{n(n-1)} \sum_{1 \leq i < j \leq n} sim(\delta_{\text{i}}, \delta_{\text{j}}), & \text{(Merge N)}
\end{cases}
\end{align}

We investigate the relationship between task performance and model kinship (see Appendix~\ref{sec:appendix-task} for the full analysis). \textbf{The results reveal strong correlations, reinforcing the view that model kinship reflects task-related differences between models.}

\subsection{Evaluation Metrics}

Let \( T \) be the set of tasks in the task group, where \( T = \{ T_1, T_2, \ldots, T_n \} \). 
Each task \( T_i \) in the set \( T \) is associated with a performance measure \( P_i \) for the LLM.
For a multi-task objective, the Average Task Performance (Avg.) \( \bar{P} \)  is calculated by:

\vspace{-10pt}

\begin{equation}
\bar{P} = \frac{1}{n}\sum_{i=1}^{n}P_{\text{i}}
\end{equation}

To evaluate the effectiveness of a single merge, we propose the merge gain metric. 
Assume we have two models $m_{pre-1}$ and $m_{pre-2}$ and their average task performance are $\bar{P}_{pre-1}$ and $\bar{P}_{pre-2}$, intuitively, we believe the $\bar{P}_{\text{merged}}$ lie around the mean of $\bar{P}_{pre-1}$ and $\bar{P}_{pre-2}$. 
The merge gain is calculated as the difference of $\bar{P}_{\text{merged}}$ from the mean value of $\bar{P}_{pre-1}$ and $\bar{P}_{pre-2}$. 
For a merging recipe with $k$ models, the merge gain is:

\vspace{-10pt}

\begin{equation}
Gain = \bar{P}_{\text{merged}} - \frac{1}{k}\sum_{i=1}^{k}\bar{P}_{\text{pre-i}}
\end{equation}

In the following analysis, we use the task group  T = \{\textit{ARC, HellaSwag, MMLU, TruthfulQA, Winogrande, GSM8K}\}. 
All models are either fine-tuned or merged from the \textit{Mistral-7B} architecture.

\begin{figure*}[ht]
\centering
\fontsize{8}{10}\selectfont
\includegraphics[width=\linewidth]{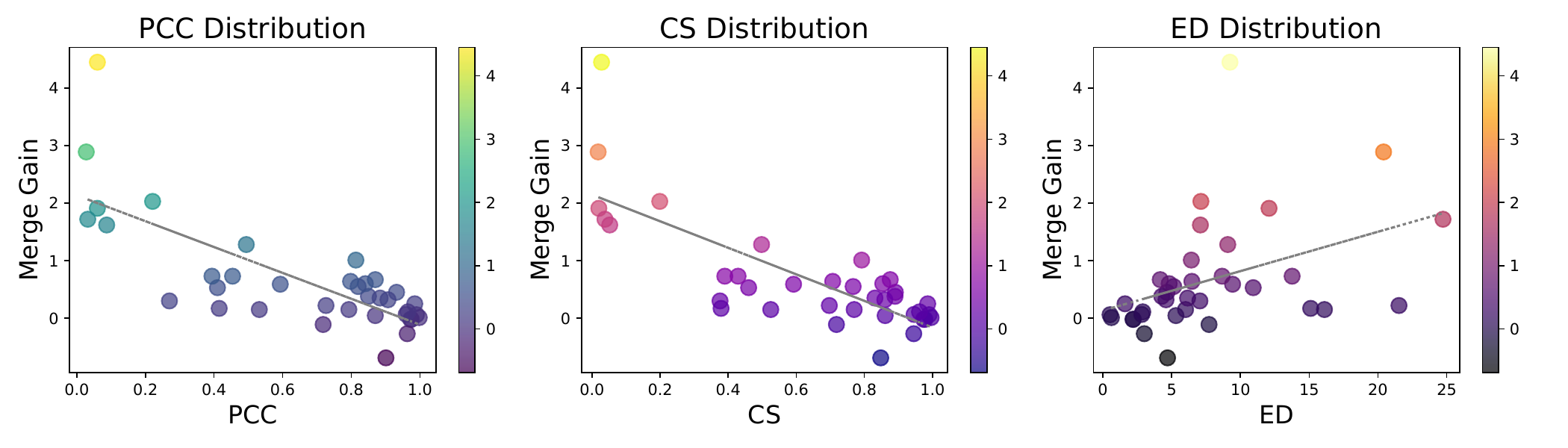}
\vspace{-20pt}
\caption{\textbf{Distribution of Sample Experiments}: Relationship Between Model Kinship (X-axis) and Merge Gain (Y-axis). Model kinships are calculated using PCC, CS, and ED.}
\label{fig:dis_results}
\vspace{-10pt}
\end{figure*}

\begin{figure*}[ht]
\centering
\includegraphics[width=\linewidth]{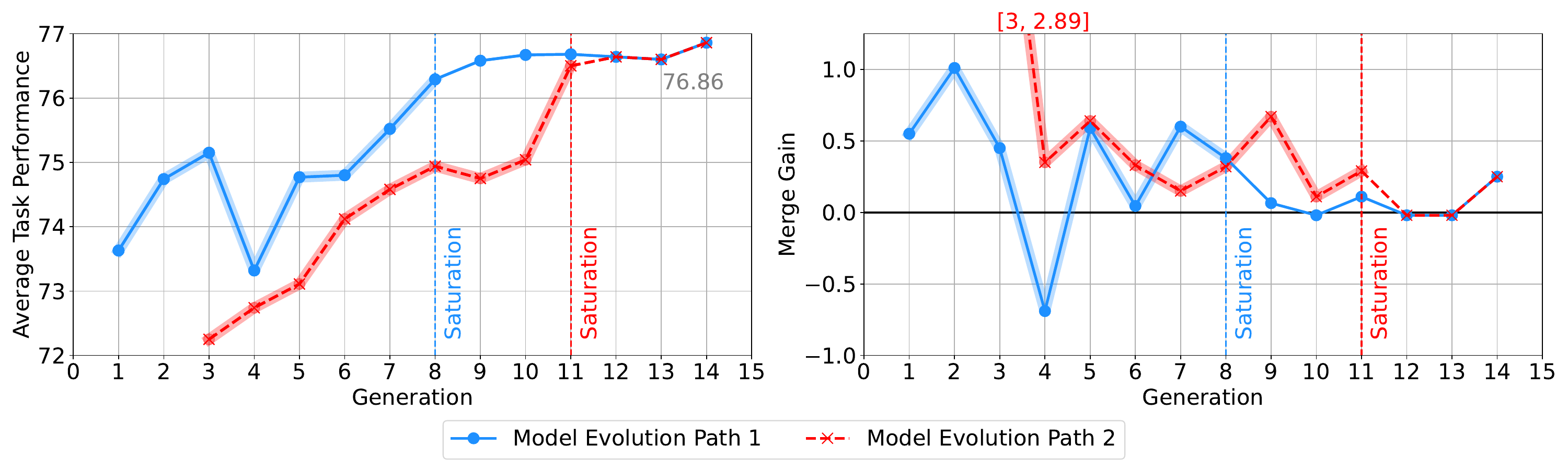}
\vspace{-15pt}
\caption{\textbf{Change in Average Task Performance and Merge Gain across the Model Evolution Process:} Paths originate from two different base models. The vertical line marks the transition to the saturation stage. Path 2 is temporally aligned with Path 1 for clarity.}
\label{fig:results1}
\end{figure*}

\subsection{Analysis of Model Kinship: Correlation and Evolution Dynamics}
\label{sec:ana2}

In this section, we analyze model kinship from two perspectives: (1) its correlation with performance gain across a broad range of open-sourced LLM merges, and (2) its dynamic along specific model evolution paths. These analyses aim to clarify the relationship between model kinship and multi-task capability improvements, as well as to identify phases of merge effectiveness. 

\subsubsection{Correlation Between Model Kinship and Performance Gain}
We begin by exploring the potential relationship between merge gain and model kinship using three similarity metrics: \textit{Pearson Correlation Coefficient (PCC)}, \textit{Cosine Similarity (CS)}, and \textit{Euclidean Distance (ED)}. 
The models used are based on the \textit{Mistral-7B} architecture \citep{jiang2023} and collected from HuggingFace, with reference to the Open LLM Leaderboard (see Appendix \ref{sec:appendix-a}).

As illustrated in Figure~\ref{fig:dis_results}, the scatter plots derived from all three metrics suggest a moderate correlation between model kinship and merge gain. Table~\ref{tab:dis_results} reports Pearson correlation values for both signed and absolute merge gains. 
While the correlations for signed gains appear relatively weak (with p-values between 0.05 and 0.1), those for absolute merge gains are comparatively stronger and show greater statistical significance. 
These observations imply that \textbf{model kinship may offer some indication of the potential magnitude of merge gains, though it appears less effective at predicting the direction of change}. While we cannot assert a causal relationship, the association provides useful insight into how kinship might relate to merge outcomes. 
In light of the comparable performance across the three metrics, we use PCC-based kinship in the remainder of our analysis for consistency.

\subsubsection{Model Kinship in Evolution Paths}

As a further exploration, we examine model kinship across independent model evolution paths to investigate potential phase patterns in the merging process. This analysis centers on the \textit{yamshadow experiment 28-7B} \citep{ye282024}, a Mistral-7B-based model that ranks among the top-performing merged models on the Open LLM Leaderboard. From its model family tree, we extract two main merging trajectories, referred to as \textbf{Path 1} and \textbf{Path 2}, for comparison.

\begin{figure*}[!ht]
\centering
\includegraphics[width=\linewidth]{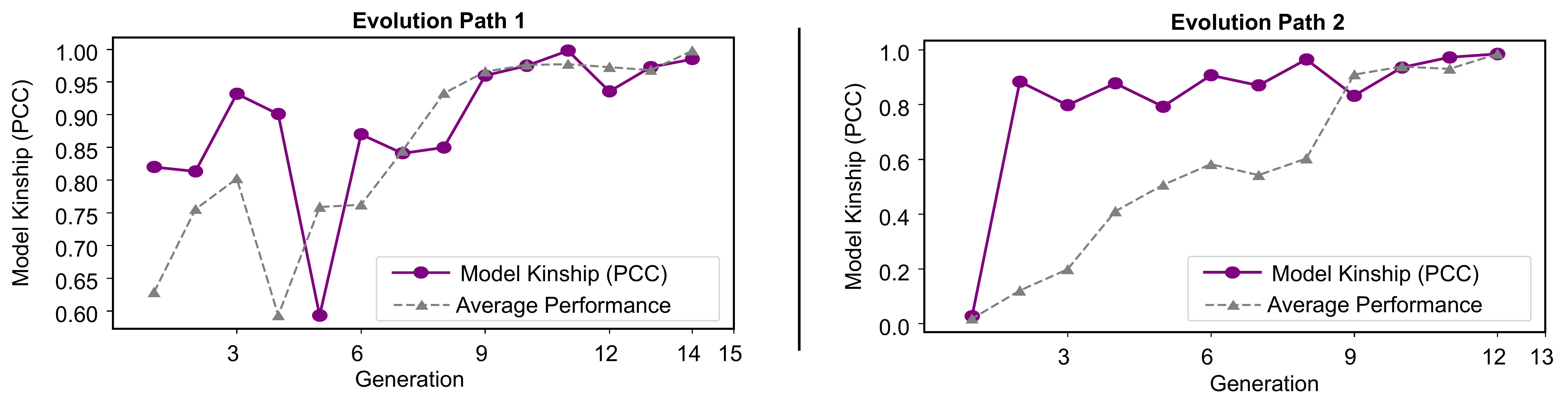}
\vspace{-20pt}
\caption{\textbf{Comparison between Model Kinship (PCC) and Average Task Performance} (normalized to the same scale).}
\label{fig:compare1}
\end{figure*}

\begin{figure*}[!ht]
    \centering
    \includegraphics[width=\linewidth]{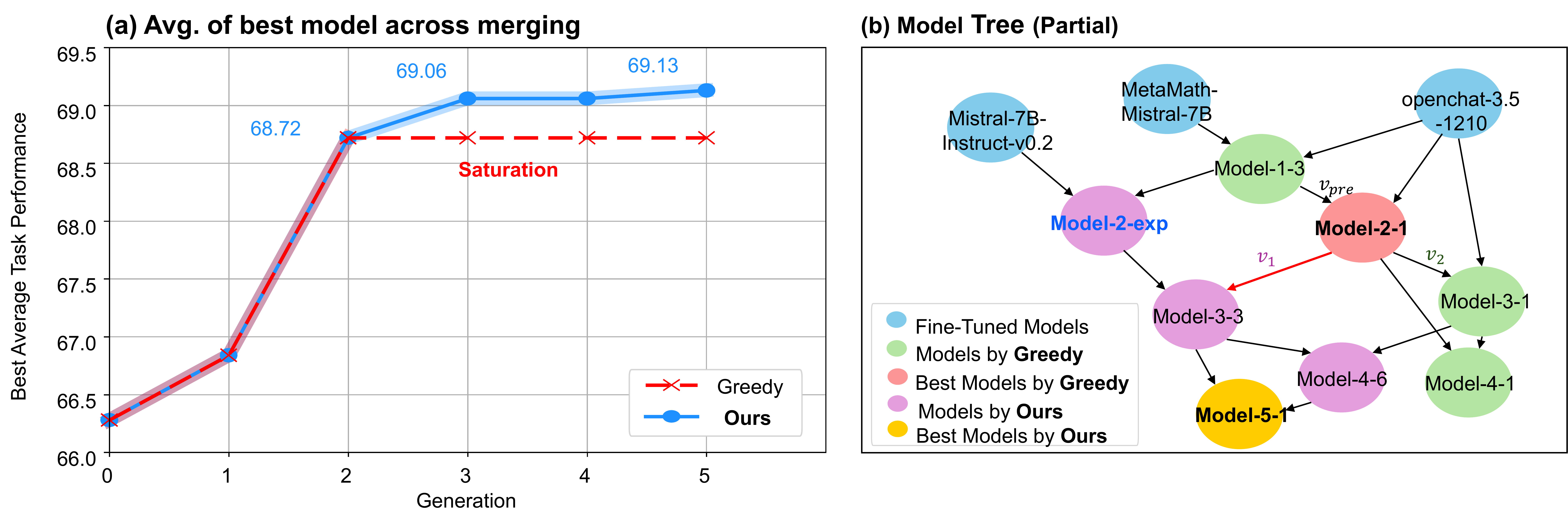}
    \vspace{-20pt}
    \caption{\textbf{Left (a)}: The comparison of task performance improvement across merging generations. The \textcolor{red}{\textbf{red curve}} (greedy strategy) saturates by generation 2, while the \textcolor{blue}{\textbf{blue curve}} (modified strategy) escapes the local optima at generation 2 and continues improving multi-task capabilities. \textbf{Right (b)}: The partial model family tree from  controlled experiments. The \textcolor{red}{\textbf{red arrow}} shows critical changes between experiment 1 and 2 in the evolution path.}
    \label{fig:results5}
\end{figure*}

Figure~\ref{fig:results1} displays the average task performance and the merge gains along the two evolution paths. The merging process exhibits two distinct phases:

\begin{itemize}
\item\textbf{Improving Stage.} Rapid performance gains and significant merge improvements, driven by active multi-task balance.

\item\textbf{Saturation Stage.} Performance stabilizes, and additional merges result in minimal or no measurable improvement.
\end{itemize}

Figure~\ref{fig:compare1} shows how the model kinship and normalized average performance change over the course of the merging process. 
Both metrics exhibit a consistent two-phase trend: an \emph{Improving Stage}, characterized by steady increases, followed by a \emph{Saturation Stage}, where growth plateaus. 
This parallel progression \textbf{highlights a potential correlation between model kinship and performance gains}, indicating that enhancements in generalization are not only concurrent with but may also be facilitated by increases in model kinship. 

\begin{table}
  \vspace{5pt}
  \centering
  \small
  \begin{tabular}{l|r|r}
    \toprule
     \textbf{Metric} & \textbf{Correlation} & \textbf{Correlation} \\
     & (Normal Value) & \textbf{(Absolute Value) }\\
    \midrule
    PCC & -0.50 & \cellcolor{gray!20} \textbf{-0.59} \\
    P-value  & 0.063 & \cellcolor{gray!20} \textbf{0.023} \\
    \midrule
    CS       & -0.45 & \cellcolor{gray!20} \textbf{-0.66} \\
    P-value  & 0.098 & \cellcolor{gray!20} \textbf{0.008} \\
    \midrule
    ED       & 0.46 & \cellcolor{gray!20} \textbf{0.67} \\
    P-value  & 0.091 & \cellcolor{gray!20} \textbf{0.007} \\
    \bottomrule
  \end{tabular}
    \caption{\textbf{Correlation} of Model Kinship based on different correlation function $sim(\cdot, \cdot)$ with Merge Gain, along with their corresponding p-values.}
    \vspace{-5pt}
  \label{tab:dis_results}
\end{table}

\begin{table*}[t]
\small
\centering
\begin{tabular}{l|cccc|l|cccc}
\toprule
\multicolumn{5}{c|}{\textbf{Greedy Strategy}} & \multicolumn{5}{c}{\textcolor{blue}{\textbf{+ Model Kinship}}} \\
\midrule
\textbf{Model} & \textbf{Avg.} & \textbf{Gain} & \textbf{$\Delta$Avg. to Top} & \textbf{Kinship} & \textbf{Model} & \textbf{Avg.} & \textbf{Gain} & \textbf{$\Delta$Avg. to Top} & \textbf{Kinship} \\ 
\midrule
\rowcolor{gray!20}
\textbf{MetaMath} & 63.72 & / & / & / & \textbf{MetaMath} & 63.72 & / & / & / \\ 
\rowcolor{gray!20}
\textbf{Instruct} & 61.82 & / & / & / & \textbf{Instruct} & 61.82 & / & / & / \\ 
\rowcolor{gray!20}
\textbf{Open-chat} & 66.28 & / & / & / & \textbf{Open-chat} & 66.28 & / & / & / \\ 
\midrule
model-1-1 & 62.17 & \textcolor{red}{-0.6} & \textcolor{red}{-4.11} & 0.01 & model-1-1 & 62.17 & \textcolor{red}{-0.6} & \textcolor{red}{-4.11} & 0.01 \\ 
model-1-2 & 64.02 & \textcolor{red}{-0.03} & \textcolor{red}{-2.26} & -0.02 & model-1-2 & 64.02 & \textcolor{red}{-0.03} & \textcolor{red}{-2.26} & -0.02 \\ 
model-1-3 & 66.84 & \textcolor{darkgreen}{+1.84} & \textcolor{darkgreen}{+0.56} & 0.05 & model-1-3 & 66.84 & \textcolor{darkgreen}{+1.84} & \textcolor{darkgreen}{+0.56} & 0.05 \\ 
\midrule
\cellcolor{blue!20}\textbf{model-2-1} & \cellcolor{blue!20}\textbf{68.72} & \cellcolor{blue!20}\textbf{\textcolor{darkgreen}{+2.16}} & \cellcolor{blue!20}\textbf{\textcolor{darkgreen}{+1.88}} & \cellcolor{blue!20}\textbf{0.93} & model-2-1 & 68.72 & \textbf{\textcolor{darkgreen}{+2.16}} & \textcolor{darkgreen}{+1.88} & 0.93 \\ 
model-2-2 & 61.47 & \textcolor{red}{-3.96} & \textcolor{red}{-7.25} & 0.57 & model-2-2 & 61.47 & \textcolor{red}{-3.96} & \textcolor{red}{-7.25} & 0.57 \\ 
model-2-3 & 61.32 & \textcolor{red}{-3.83} & \textcolor{red}{-7.40} & 0.58 & model-2-3 & 61.32 & \textcolor{red}{-3.83} & \textcolor{red}{-7.40} & 0.58 \\ 
\midrule
model-3-1 & 68.59 & \textcolor{darkgreen}{+1.09} & \textcolor{red}{-0.13} & 0.95 & model-3-2 & 67.74 & \textcolor{darkgreen}{+1.09} & \textcolor{red}{-0.98} & 0.93 \\ 
model-3-2 & 67.74 & \textcolor{red}{-0.04} & \textcolor{red}{-0.98} & 0.93 & model-3-3 & 69.06 & \textcolor{darkgreen}{+0.74} & \textcolor{darkgreen}{+0.34} & 0.24 \\ 
 & - & - & - & - & model-3-4 & 68.61 & \textcolor{darkgreen}{+1.13} & \textcolor{red}{-0.11} & 0.32 \\ 
\midrule
model-4-1 & 68.51 & \textcolor{red}{-0.14} & \textcolor{red}{-0.08} & 0.98 & model-4-4 & 68.75 & \textcolor{red}{-0.14} & \textcolor{red}{-0.31} & 0.54 \\ 
model-4-2 & 68.04 & \textcolor{red}{-0.19} & \textcolor{red}{-0.67} & 0.98 & model-4-5 & 68.39 & \textcolor{red}{-0.27} & \textcolor{red}{-0.36} & 0.66 \\ 
model-4-3 & 68.53 & \textcolor{darkgreen}{+0.37} & \textcolor{red}{-0.06} & 0.94 & model-4-6 & 69.03 & \textcolor{darkgreen}{+0.15} & \textcolor{darkgreen}{+0.42} & 0.52 \\ 
\midrule
 & - & - & - & - & \cellcolor{blue!20}\textbf{model-5-1} & \cellcolor{blue!20}\textbf{69.13} & \cellcolor{blue!20}\textbf{\textcolor{darkgreen}{+0.04}} & \cellcolor{blue!20}\textcolor{darkgreen}{+0.10} & \cellcolor{blue!20}\textbf{0.65} \\ 
 & - & - & - & - & model-5-2 & 68.98 & \textcolor{darkgreen}{+0.07} & \textcolor{red}{-0.15} & 0.65 \\ 
 & - & - & - & - & model-5-3 & 68.63 & \textcolor{red}{-0.37} & \textcolor{red}{-0.50} & 0.98 \\ 
\bottomrule
\end{tabular}
\vspace{-8pt}
\caption{\textbf{Results of merging experiments} comparing the vanilla greedy strategy and our proposed approach. The first three models serve as the foundation models in both experiments. \textbf{Note}: The model kinship experiment was terminated at generation 5, as it has already outperformed the greedy strategy by that point.}
\label{tab:main}
\vspace{-10pt}
\end{table*}

To move beyond individual Evolution Paths, we further investigate how model kinship develops across different stages of the merging process, thereby extending our analysis to the broader evolutionary landscape (refer to  Appendix~\ref{sec:appendix-stage}).
Our findings reveal that the highest-performing models maintain strong mutual kinship. However, this close relatedness can also induce a stagnation phase, where the lack of diversity limits further performance improvements.

\subsection{Discussion}

Considering all results that we observed, this analysis provides two main insights for the application of model kinship:

\begin{itemize}
    \item \textbf{High kinship merges may lead to performance stagnation, similar to biological inbreeding, where excessive similarity limits the ability to adjust.}
    \item \textbf{Low kinship merges involve risk, but can result in greater gains, potentially enabling escape from local optima caused by the greedy strategy.}
\end{itemize}

\section{Using Model Kinship to Improve Model Merging}
\label{sec:method}
Building on the insights from \cref{sec:emprical_results}, we explore how model kinship can be leveraged to improve the model merging process.
Our main experiment centers on the \textit{Mistral-7B} model, with detailed results presented in the main text. To further evaluate the generalization of our approach, we also conduct two supplementary experiments: one based on \textit{Llama-2} (see Appendix~\ref{sec:appendix-full}) and another on a distinct task set to test robustness across different evaluation settings.
\textbf{Our results indicate that while greedy strategy focuses on short-term gains, it can lead to suboptimal outcomes. }
By integrating model kinship, we can help the strategy avoid local optima and gain further improvements. 

\subsection{Main Experiment Setup}
\label{sec:exp}

For the main experiments, we follow the same settings as described in \cref{sec:model_evolution}, including the use of the three fine-tuned \textit{Mistral-7B} variants and the evaluation on Winogrande, GSM8k, and TruthfulQA. We adopt the Top-$k$ Greedy Merging strategy as the baseline iterative merging strategy\footnote{We provide a Google Colab \href{https://colab.research.google.com/drive/141VAI89emgSIcwkswATEXSEENoAMywTO?usp=sharing}{Notebook}.}.

\paragraph{Top $k$ Greedy Merging with Model Kinship.}
We propose an enhanced merging strategy that augments the original greedy approach with an additional exploration step guided by model kinship (highlighted in \textcolor{blue}{blue} in Algorithm~\ref{alg:greedy_merge}).
This approach aims to merge the best-performing model with the model that has the most distinct kinship, in order to discover potentially better solutions. 
In Figure \ref{fig:results5} (b), models generated by our strategy are marked in purple, while the best-performing models are marked in yellow.

\subsection{Results and Discussion}
\label{local-optima-2}

\begin{wrapfigure}{r}{0.2\textwidth}
    \vspace{-10pt}
    \centering
    \includegraphics[width=\linewidth]{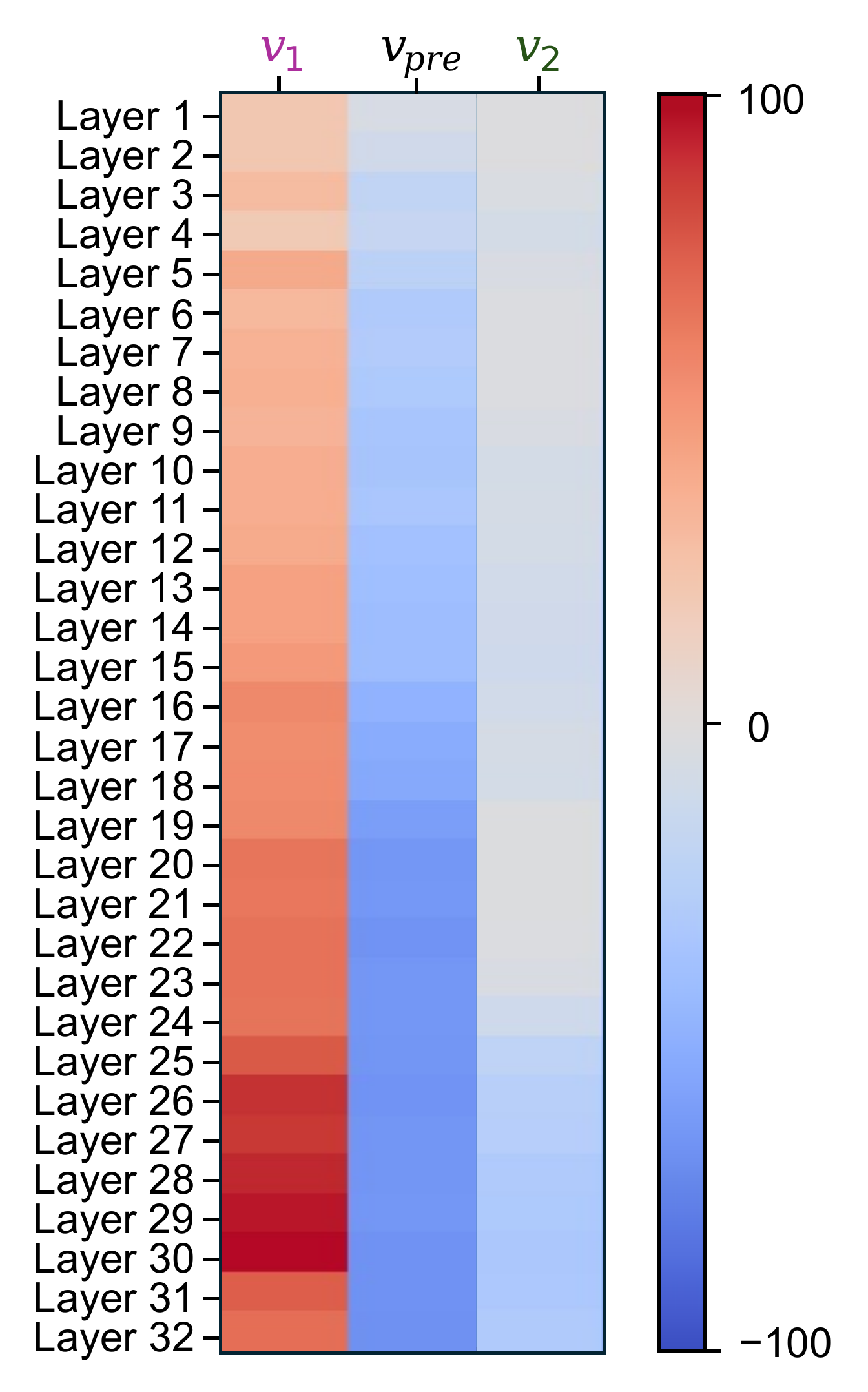}
    \caption{Weight Change between two Evolution Paths.}
    \vspace{-10pt}
    \label{fig:heatmap}
\end{wrapfigure}

Figure \ref{fig:results5} (a) illustrates the improvements in top average task performance across merging generations.
Table \ref{tab:main} provides model average task performance, merge gain, and model kinship for each generation, comparing the original greedy merging strategy with our kinship-based method.
While both strategies achieve the multi-task objective, the \textit{vanilla greedy strategy} ceases to improve after Generation 2, plateauing at an average task performance of \textbf{68.72}. 
In contrast, strategy utilizing model kinship escapes the local optima (Model-2-1) and continues to improve, reaching \textbf{69.13} by Generation 5.

\paragraph{Merging Models with Low Kinship can Boost Exploration.}
Figure \ref{fig:results5} (b) shows the main branch of the model family tree. 
To explore how low-kinship merges help escape local optima during saturation, we analyze weight changes: $v_1$ (from \textit{Model-2-1} to \textit{Model-3-1}) and $v_2$ (from \textit{Model-2-1} to \textit{Model-3-3}), with $v_{pre}$ (from \textit{Model-1-3} to \textit{Model-2-1}) as baseline. Figure~\ref{fig:heatmap} shows that merging with the exploration model ($v_2$) \textbf{produces significant, novel weight changes}, while $v_1$ shows minimal change due to the high similarity between \textit{openchat-3.5} and \textit{Model-2-1}.

\paragraph{Early Stopping at High Kinship can Improve Efficiency.}  

Iterative model merging can be resource intensive. In our main experiments, a greedy strategy saturated after \textbf{2/4} merges with no further gains. Looking back at community experiments, \textbf{5/14} merges in path 1 averaged only \textbf{0.57} improvement, and \textbf{3/12} merges in path 2 averaged \textbf{0.36}. A high kinship score (PCC > 0.9) among top models may indicate convergence. \textbf{Stopping merges early at high kinship generation saves ~30\% time} with negligible performance loss.

\section{Related work}

Weight averaging, a widely used technique in model merging, traces its origins to \citet{utans1996weight}.
Since the 2010s, weight averaging has been widely applied in deep neural networks, notably for combining checkpoints to improve training stability and performance.
\citep{Nagarajan2019,tarvainen2017mean,Izmailov2018AveragingWL,li2023deep,stoica2023zipit,padmanabhan2023gemel,jang2023personalized}, leveraging task-specific information \citep{li2023, ilharco2022, Izmailov2018AveragingWL, joshuasmith2017}, and parallel training of large language models (LLMs) \citep{li2022branch}. 
Discovery of Linear Mode Connectivity (LMC) \citep{Garipov2018,Frankle2020,entezari2022the} further expands the use of weight averaging in fusing fine-tuned models through averaging methods \citep{what2020,wortsman2022}. 
Further studies have explored optimizable weights for merging, such as Fisher-Merging \citep{matena2022}, RegMean \citep{jin2022}, AdaMerging \citep{Yang2024}, MaTS \citep{Tam2023}. 
\citet{ilharco2023} introduce task vectors, representing the weight difference between a fine-tuned model and its base. 
Further research on parameter interference led to TIES \citep{yadav2023}, which preserves important weights and reduces sign conflicts, and DARE \citep{yu2024}, which prevents interference by randomly dropping weights. 
The Model Breadcrumbs \citep{davari2024model} show that the removal of outliers in parameters can reduce noise in model merging. 
Merging models with different initializations requires additional considerations.
Common methods exploit the permutation symmetry of neural networks \citep{ainsworth2023git, Tatro2020, SinghJ20, Guerrero2023}, using alignment techniques to mitigate the interpolation barrier \citep{Xu2024, Navon2024}. 
While weight averaging cannot be applied to models with different architectures, it can still be used to enhance feasible fusion methods. 
Recent work like FuseChat \citep{fusechat2024} integrates it with Knowledge Fusion \citep{wan2024} to enable novel fusion approaches.
\citet{li2025pma} further demonstrate that model merging can stabilize training and serve as a low-cost simulator for annealed performance during pretraining, enabling checkpoint reuse and faster validation.

Recently, there have been some works exploring ``model evolution''.
\citet{don-yehiya2023cold} propose the CoLD Fusion method, showing that iterative fusion can create effective multi-task models. 
\citet{ye282024} develop a tool to automatically merge models on HuggingFace.
\citet{akiba2024} introduce Evolutionary Model Merge, employing evolutionary techniques to optimize model combinations.
\citet{tang2025} present a continual model merging method that enables training-free projection-based merging of models as they arrive sequentially, improving efficiency and task retention without retraining.

\section{Conclusion}
We propose \textbf{iterative model merging} as a framework for evolving large language models through strategic, iterative merges that yield consistent gains in generalization and task performance. 
To support this framework, we introduce \textit{model kinship}, a metric that guides merge candidate selection and explains both performance gains and stagnation during merging. 
Building on this, we propose \textbf{Top-$k$ Greedy Merging with Model Kinship}, a strategy that uses kinship to escape local optima and achieve further improvements. 
Kinship also serves as an early stopping signal by detecting convergence and reducing redundant computation.

In a broad sense, our work explores how models can achieve autonomous evolution through model merging. 
Model merging can, to some extent, be likened to biological hybridization.
Biological organisms have undergone billions of years of evolution to reach their current state.
However, how silicon-based intelligence, represented by LLMs, evolves remains an unresolved mystery. 
We aspire that this work offer guidance and insights for the future merging and evolution of LLMs.

\section*{Limitations}
This study has several limitations:  
\textit{a)} Our experiments are limited to two model architectures. It remains unclear whether our metric and method generalize to other architectures, such as \textit{Mamba} \citep{mamba}. In addition, the scalability of tasks and candidate models requires further evaluation to assess computational costs in diverse settings.  
\textit{b)} The analysis relies on open-source data from the Open Leaderboard. As this dataset is community-generated, it may be affected by noise or user bias.  
\textit{c)} We have not fully explored alternative correlation metrics for measuring model kinship. Other metrics may yield stronger or more consistent performance.  
\textit{d)} Our demonstration of model kinship is empirical. A more rigorous theoretical framework, such as the assumptions outlined in Appendix \ref{sec:appendix-c}, is needed to explain model evolution and kinship in greater depth.  
\textit{e)} While model kinship currently guides merging and improves performance, it does not support sustained evolution. Future progress may require incorporating environmental feedback, reward models \citep{silver2021reward}, and novel architectures.

\section*{Acknowledgments}
This work was supported by the National Natural Science Foundation of China (No. 62576307), Ningbo Natural Science Foundation (2024J020), the Ministry of Education, Singapore, under the Academic Research Fund Tier 1 (FY2023) (Grant A-8001996-00-00), 
the Academic Research Fund Tier 1 (FY2025) (Grant T1 251RES2507),
Tencent AI Lab Rhino-Bird Focused Research Program (RBFR2024003), 
and  
Information Technology Center and State Key Lab of CAD\&CG, Zhejiang University.

\bibliography{custom}

\clearpage

\appendix

\section{Details of Experiments in Main Sections}
\label{sec:appendix-a}

This section provides comprehensive details on the models used in the analysis of community experiments. 
The open merged models from these experiments are accessible through the Hugging Face Hub\footnote{\url{https://huggingface.co/datasets}}. The evaluation results can be accessed in the Openleaderboard\footnote{\url{https://huggingface.co/spaces/open-llm-leaderboard-old/open_llm_leaderboards}} The following tables cover two primary aspects: 

\begin{itemize}
    \item \textbf{(1)} Information on the selected model family trees for two distinct evolution paths, along with detailed analysis results for each merge.
    \item \textbf{(2)} A summary of the merge experiments conducted for distribution analysis.
\end{itemize}

\subsection{Selecting the Evolution Path}

The evolution paths are selected using a structured process, focusing on identifying key sequences within the model family trees. The steps were as follows:
\begin{itemize}
    \item \textbf{Model Family Tree Construction}: The complete model family tree is constructed by referencing model card details for each model involved.
    \item \textbf{Branch Identification}: We identified the two longest branches within each tree, representing significant sequences of model merging.
    \item \textbf{Performance and Kinship Evaluation}: These branches were analyzed for changes in merging performance, particularly focusing on shifts in multi-task capabilities and model kinship metrics.
\end{itemize}

Table \ref{tab:models1} and \ref{tab:models2} present detailed information on the sequential merging process. The second and third columns record the foundational models involved in each merge, while the final column lists the resulting merged models. 

\subsection{Additional Results in Analysis}

Table \ref{tab:evolution1} and Table \ref{tab:evolution2} present detailed analysis results that are not reported in the main paper. These include Average Task Performance (ATP), merge gains, and model kinship values, which are computed using Pearson Correlation coefficient, Cosine Similarity, and Euclidean Distance for each merge.

Table \ref{tab:distribution} presents all merge experiments contributing to the distribution analysis. The selection of sample experiments adheres to two rules: \textbf{(1)}  Samples are evenly chosen across average task performance values ranging from \textit{0.7} to \textit{0.7686} (the average task performance of the best 7B merged model) to accurately reflect the full scope of model evolution. \textbf{(2)}  The experiments involve merges of two foundation models, as including multiple models introduces excessive noise.

\subsection{Details of Datasets Selection}
\label{sec:dataset}

In the main experiments, we utilize three task-specific benchmark datasets—Winogrande, GSM8k, and TruthfulQA—to evaluate the distinct strengths of the models. These datasets assess the following capabilities:

\begin{itemize}
\item \textbf{Winogrande}: Evaluates the model's commonsense reasoning.
\item \textbf{GSM8k}: Measures the model's mathematical reasoning.
\item \textbf{TruthfulQA}: Assesses the model's ability to identify and address human falsehoods.
\end{itemize}

\clearpage

\begin{table*}[ht]
  \centering
  \small
  \begin{tabular}{clll}
    \toprule
    \textbf{Gen} & \textbf{Model-1} & \textbf{Model-2} & \textbf{Model-Merged} \\
    \midrule
    1  & Marcoroni-7B-v3 &  Mistral-7B-Merge-14-v0.1 & distilabeled-Marcoro14-7B-slerp\\
    2  & distilabeled-Marcoro14-7B & UNA-TheBeagle-7b-v1 & Beagle14-7B \\
    3  & NeuralBeagle14-7B & Turdus & TurdusBeagle-7B    \\
    4  & TurdusBeagle-7B & FernandoGPT-v1 & StrangeMerges\_9-7B-dare\_ties   \\
    5  & StrangeMerges\_9-7B-dare\_ties & MBX-7B-v3  & StrangeMerges\_10-7B-slerp   \\
    6  & StrangeMerges\_10-7B-slerp  & NeuralBeagle14-7B  & StrangeMerges\_11-7B-slerp    \\
    7  & StrangeMerges\_11-7B-slerp  & MBX-7B-v3  & StrangeMerges\_20-7B-slerp    \\
    8  & StrangeMerges\_20-7B-slerp  & NeuTrixOmniBe-7B-model & StrangeMerges\_21-7B-slerp \\
    9  & StrangeMerges\_21-7B-slerp  & Experiment26 & StrangeMerges\_30-7B-slerp     \\
    10  & StrangeMerges\_30-7B-slerp  & Experiment24 & StrangeMerges\_31-7B-slerp  \\
    11  & StrangeMerges\_31-7B-slerp  & Experiment28 & StrangeMerges\_32-7B-slerp \\
    12  & StrangeMerges\_32-7B-slerp  & ... &  shadow-clown-7B-slerp \\
    13  & shadow-clown-7B-slerp  & yam-jom-7B  & YamShadow-7B   \\
    14  & YamShadow-7B  & Experiment28 & YamshadowExperiment28-7B  \\
    \bottomrule
  \end{tabular}
    \caption{Model Family tree of evolution Path 1.}
  \label{tab:models1}
\end{table*}

\begin{table*}[ht]
  \centering
  \small
  \begin{tabular}{clll}
    \toprule
    \textbf{Gen} & \textbf{Model-1} & \textbf{Model-2} & \textbf{Model-Merged} \\
    \midrule
    1 & neural-chat-7b-v3-3 & openchat-3.5-1210 & CatPPT-base \\
    2 & Marcoroni-7B-v3 & CatPPT-base & CatMacaroni-Slerp \\
    3 & LeoScorpius-7B & CatMacaroni-Slerp & SamirGPT-v1 \\
    4  & SamirGPT-v1 & ... & Daredevil-7B \\
    5  & NeuralBeagle14-7B & NeuralDaredevil-7B & DareBeagle-7B \\
    6  & Turdus & DareBeagle-7B & TurdusDareBeagle-7B \\
    7  & MarcMistral-7B & TurdusDareBeagle-7B & MarcDareBeagle-7B \\
    8  & MarcBeagle-7B & MarcDareBeagle-7B & MBX-7B \\
    9  & MBX-7B & ... & pastiche-crown-clown-7b-dare \\
    10  & pastiche-crown-clown-7b-dare & ... & shadow-clown-7B-slerp \\
    11  & yam-jom-7B & shadow-clown-7B-slerp & YamShadow-7B \\
    12  & Experiment28-7B & YamShadow-7B & YamshadowExperiment28-7B \\
    \bottomrule
  \end{tabular}
      \caption{Model Family tree of evolution Path 2.}
  \label{tab:models2}
\end{table*}

\begin{table*}[ht]
  \centering
  \begin{tabular}{ccccccc}
    \toprule
    \textbf{Gen} & \textbf{Model-Merged}& \textbf{ATP} & \textbf{Gain} & \textbf{PCC} & \textbf{CS} & \textbf{ED} \\
    \midrule
    1 & distilabeled-Marcoro14-7B-slerp   & 73.63 & 0.55   & 0.82  & 0.76 & 5.15 \\
    2 & Beagle14-7B  & 74.74 & 1.01   & 0.81& 0.79 & 6.43 \\
    3  & StrangeMerges\_9-7B-dare\_ties  & 75.15 & 0.45   & 0.93 & 0.89 & 4.66 \\
    4  & StrangeMerges\_9-7B-dare\_ties  & 73.32 & -0.69  & 0.90 & 0.84 & 4.70 \\
    5   & StrangeMerges\_10-7B-slerp   & 74.77 & 0.59   & 0.59 & 0.59 & 9.43 \\
    6  & StrangeMerges\_11-7B-slerp   & 74.8  & 0.045  & 0.87  & 0.86 & 5.31 \\
    7  & StrangeMerges\_20-7B-slerp   & 75.52 & 0.6    & 0.84 & 0.85 & 4.82 \\
    8  & StrangeMerges\_21-7B-slerp  & 76.29 & 0.38   & 0.85  & 0.89 & 4.28 \\
    9 & StrangeMerges\_30-7B-slerp  & 76.58 & 0.065  & 0.96  & 0.94 & 2.83 \\
    10 & StrangeMerges\_31-7B-slerp & 76.67 & -0.02  & 0.97 & 0.97 & 2.21 \\
    11  & StrangeMerges\_32-7B-slerp  & 76.68 & 0.11   & 0.99 & 0.99 & 0.62 \\
    12 &  shadow-clown-7B-slerp  & 76.64 & -0.02  & 0.93& 0.94& 2.49 \\
    13 & YamShadow-7B  & 76.6  & -0.02  & 0.97 & 0.97 & 2.19 \\
    14  & YamshadowExperiment28-7B  & 76.86 & 0.25   & 0.98 & 0.98 & 1.61 \\
    \bottomrule
  \end{tabular}
    \caption{Summary of Path 1 Results.}
  \label{tab:evolution1}
\end{table*}

\begin{table*}[ht]
  \small
  \centering
  \begin{tabular}{ccccccc}
    \toprule
    \textbf{Gen}& \textbf{Model-Merged}  & \textbf{ATP} & \textbf{Gain} & \textbf{PCC} & \textbf{CS} & \textbf{ED} \\
    \midrule
    1   & CatPPT-base & 72.25 & 2.89   & 0.02         & 0.01         & 20.41 \\
    2 & CatMacaroni-Slerp   & 72.74 & 0.35   & 0.88          & 0.83         & 6.16  \\
    3 & SamirGPT-v1   & 73.11 & 0.64   & 0.79          & 0.70         & 6.47  \\
    4 & Daredevil-7B  & 74.12 & 0.33   & 0.87   & 0.83  & 4.81 \\
    5 & DareBeagle-7B  & 74.58 & 0.15   & 0.79          & 0.77         & 6.01  \\
    6  & TurdusDareBeagle-7B  & 74.94 & 0.32   & 0.90          & 0.86         & 4.59  \\
    7  & MarcDareBeagle-7B   & 74.75 & 0.67   & 0.87          & 0.87         & 4.17  \\
    8  & MBX-7B & 75.04 & 0.11   & 0.96          & 0.96         & 2.90  \\
    9 & pastiche-crown-clown-7b-dare  & 76.50 & 0.29   & 0.83    & 0.84& 5.38 \\
    10 & shadow-clown-7B-slerp  & 76.64 & -0.02  & 0.93    & 0.94   & 2.49 \\
    11 & YamShadow-7B & 76.60 & -0.02  & 0.97         & 0.97         & 2.19  \\
    12& YamshadowExperiment28-7B  & 76.86 & 0.25   & 0.98          & 0.98         & 1.61  \\
    \bottomrule
  \end{tabular}
\caption{Summary of Path 2 Results.}
  \label{tab:evolution2}
\end{table*}

\begin{table*}[ht]
  \centering
  \begin{tabular}{ll|c}
    \toprule
    \textbf{Model 1} & \textbf{Model 2} & \textbf{Merge Gain} \\
    \midrule
    Multi\_verse\_model-7B & Experiment26-7B & 0.06 \\
    M7-7b & StrangeMerges\_32-7B-slerp & -0.03 \\
    Ognoexperiment27 & Multi\_verse\_model-7B & 0.03 \\
    YamShadow-7B & Experiment28 & 0.25 \\
    shadow-clown-7B-slerp & yam-jom-7B & -0.02 \\
    StrangeMerges\_21-7B-slerp & Experiment26 & 0.06 \\
    StrangeMerges\_31-7B-slerp & Experiment28 & 0.11 \\
    NeuralBeagle14-7B & Turdus & 0.45 \\
    DareBeagle-7B & Turdus & 0.32 \\
    TurdusBeagle-7B & FernandoGPT-v1 & -0.69 \\
    StrangeMerges\_10-7B-slerp & NeuralBeagle14-7B & 0.04 \\
    TurdusDareBeagle-7B & MarcMistral-7B & 0.67 \\
    StrangeMerges\_20-7B-slerp & NeuTrixOmniBe-7B-model-remix & 0.38 \\
    StrangeMerges\_11-7B-slerp & MBX-7B-v3 & 0.6 \\
    Marcoroni-7B-v3 & Mistral-7B-Merge-14-v0.1 & 0.55 \\
    distilabeled-Marcoro14-7B-slerp & UNA-TheBeagle-7b-v1 & 1.01 \\
    UNA-TheBeagle-7b-v1 & distilabeled-Marcoro14-7B-slerp & 0.89 \\
    CatPPT-base & Marcoroni-7B-v3 & 0.35 \\
    CatMacaroni-Slerp & LeoScorpius-7B & 0.64 \\
    NeuralDaredevil-7B & NeuralBeagle14-7B & 0.15 \\
    StrangeMerges\_9-7B-dare\_ties & MBX-7B-v3 & 0.59 \\
    mistral-ft-optimized-1218 & NeuralHerems-Mistral-2.5-7B & -0.85 \\
    neural-chat-7b-v3-2 & OpenHermes-2.5-Mistral-7B & 1.91 \\
    StrangeMerges\_30-7B-slerp & Experiment24 & -0.02 \\
    openchat-3.5-1210 & neural-chat-7b-v3-3 & 2.89 \\
    MultiverseEx26-7B-slerp & CalmExperiment-7B-slerp & -0.09 \\
    CapybaraMarcoroni-7B & DistilHermes-2.5-Mistral-7B & 0.47 \\
    Multi\_verse\_model-7B & Calme-7B-Instruct-v0.9 & 0.04 \\
    StrangeMerges\_16-7B-slerp & coven\_7b\_128k\_orpo\_alpha & -0.35 \\
    Kunoichi-DPO-v2-7B & AlphaMonarch-7B & -1.05 \\
    StrangeMerges\_15-7B-slerp & Kunoichi-7B & 0.39 \\
    Mistral-T5-7B-v1 & Marcoroni-neural-chat-7B-v2 & -0.18 \\
    Marcoro14-7B-slerp & mistral-ft-optimized-1218 & -0.61 \\
    mistral-ft-optimized-1218 & NeuralHermes-2.5-Mistral-7B & -0.85 \\
    MarcDareBeagle-7B & MarcBeagle-7B & -0.07 \\
    MetaMath-Mistral-7B & Tulpar-7b-v2 & -0.29 \\
    YugoGPT & AlphaMonarch-7B & -5.96 \\
    \bottomrule
  \end{tabular}
    \caption{\textbf{All Sample Experiments} used in distribution analysis.}
  \label{tab:distribution}
\end{table*}

\clearpage

\section{Full Evaluation Results of Main Experiments and Additional Experiments}
\label{sec:appendix-full}

\subsection{Main Mistral-7B Experiments}
Table \ref{tab:full1} provides a detailed evaluation of the main experiments, including the results for the exploration models and their performance on specific tasks. The model kinship experiment is terminated early at generation 5, as a more promising evolution path is subsequently identified.

\subsection{Additional Experiments}
To assess the generalization of our strategy across different model architectures and task sets, we conduct two additional experiments.

\subsubsection{Mistral-7B Experiments with a Different Task Set}
We perform further evaluations using Mistral-7B, based on three distinct foundation models: \textbf{\textit{MistralHermes-CodePro-7B-v1}}, \textbf{\textit{metamath-mistral-7b}}, and \textbf{\textit{open-chat-3.5-1210}}. These models are assessed on the HumanEval, GSM8k, and TruthfulQA benchmarks. The model kinship-based merging process is terminated early at generation 3, as a more promising evolution trajectory is identified.

In this task setting, model kinship-guided exploration successfully discovers models (e.g., Child3-3) that significantly outperform their respective initial baselines.

\subsubsection{LLaMA-2-8B Experiments}
We further evaluate our strategy on LLaMA-2-8B using three task-specific fine-tuned models.
Table~\ref{tab:full2} summarizes the results of these additional experiments. The model kinship-based merging process is terminated early at generation 6 upon the discovery of a more favorable evolutionary path.

Consistent with the results from Mistral-7B, model evolution guided by model kinship continues to facilitate performance improvements beyond the capabilities of the original models.

\clearpage

\begin{table*}[ht]
\centering
\begin{tabular}{l|ccc|c|c}
\toprule
\textbf{Model} & \textbf{TruthfulQA} & \textbf{Winogrande} & \textbf{GSM8K} & \textbf{Avg.} & \textbf{Model Kinship} \\ 
\midrule
MetaMath & 44.89 & 75.77 & 70.51 & 63.72 & / \\
Instruct & 68.26 & 77.19 & 40.03 & 61.82 & / \\
Open-chat & 52.15 & 80.74 & 65.96 & 66.28 & /\\
\midrule
model-1-1-greedy & 52.51 & 76.16 & 57.85 & 62.17 & 0.01 \\
model-1-2-greedy & 58.04 & 76.32 & 57.72 & 64.02 & -0.02 \\
model-1-3-greedy & 48.96 & 78.69 & 72.86 & 66.84 & 0.05 \\
\midrule
model-2-1-greedy & 50.94 & 80.11 & 75.13 & \textbf{\textcolor{red}{68.72}} & 0.93 \\
model-2-2-greedy & 49.78 & 78.93 & 55.72 & 61.47 & 0.57 \\
model-2-3-greedy & 52.36 & 78.61 & 52.99 & 61.32 & 0.58 \\
\textbf{model-2-exp} & 61.01 & 79.56 & 63.76 & 68.11 & -0.02 \\ 
\midrule
model-3-1-greedy & 51.95 & 80.51 & 73.31 & 68.59 & 0.95 \\
model-3-2-greedy & 49.96 & 79.72 & 73.54 & 67.74 & 0.93 \\
model-3-3 & 56.95 & 80.25 & 70.00 & 69.06 & 0.24 \\
model-3-4 & 54.38 & 78.45 & 73.01 & 68.61 & 0.32 \\
\textbf{model-3-exp} & 54.13 & 78.69 & 71.65 & 68.15 & 0.03 \\ 
\midrule
model-4-1-greedy & 50.82 & 80.11 & 74.60 & 68.51 & 0.98 \\
model-4-2-greedy & 50.36 & 79.47 & 74.31 & 68.04 & 0.98 \\
model-4-3-greedy & 51.04 & 79.72 & 74.83 & 68.53 & 0.94 \\
model-4-4 & 53.31 & 79.40 & 73.54 & 68.75 & 0.54 \\
model-4-5 & 52.48 & 79.01 & 73.68 & 68.39 & 0.66 \\
model-4-6 & 53.69 & 79.72 & 73.69 & 69.03 & 0.52 \\
\textbf{model-4-exp} & 55.16 & 78.53 & 71.80 & 68.49 & 0.48 \\ 
\midrule
model-5-1 & 54.85 & 79.37 & 73.31 & \textbf{\textcolor{darkgreen}{69.13}} & 0.65 \\
model-5-2 & 54.78 & 79.40 & 72.86 & 68.98 & 0.65 \\
model-5-3 & 53.49 & 79.24 & 73.16 & 68.63 & 0.98 \\
\textbf{model-5-exp} & 52.98 & 79.32 & 72.78 & 68.36 & 0.59 \\ 
\bottomrule
\end{tabular}
\caption{Evaluation Results of Main Experiments of Mistral-7B.}
\label{tab:full1}
\end{table*}

\begin{table*}[ht]
\centering
\begin{tabular}{l|ccc|c|c}
\toprule
\textbf{Model} & \textbf{TruthfulQA} & \textbf{GSM8K} & \textbf{HumanEval} & \textbf{Avg.} & \textbf{Model Kinship} \\
\midrule
MetaMath & 44.89 & 70.51 & 17.68 & 44.36 & / \\
Openchat-3.5-1210 & 52.15 & 65.96 & 2.44 & 40.18 & / \\
MistralHermes-CodePro-7B-v1 & 49.67 & 60.88 & 22.56 & 44.37 & / \\
\midrule
child1-1-greedy & 51.87 & 69.60 & 15.80 & 45.76 & 0.19 \\
child1-2-greedy & 48.04 & 72.78 & 9.15 & 43.32 & 0.08 \\
child1-3-greedy & 48.96 & 72.86 & 18.29 & 46.70 & 0.05 \\
\midrule
child2-1-greedy & 50.24 & 71.72 & 12.20 & 44.72 & 0.15 \\
child2-2-greedy & 50.88 & 73.24 & 7.32 & 43.81 & 0.92 \\
child2-3-greedy & 51.15 & 67.32 & 19.51 & 45.99 & 0.34 \\
child2-4-greedy & 50.05 & 72.33 & 4.88 & 42.42 & 0.21 \\
\textbf{child2-exp} & 50.33 & 71.11 & 18.90 & 46.78 & 0.21 \\
\midrule
child3-1-greedy & 51.47 & 69.22 & 21.34 & \textbf{\textcolor{red}{47.34}} & 0.73 \\
child3-2-greedy & 50.71 & 72.40 & 9.15 & 44.09 & 0.82 \\
child3-3 & 49.69 & 74.37 & 21.34 & \textbf{\textcolor{darkgreen}{48.47}} & 0.82 \\
child3-4 & 50.57 & 69.75 & 17.68 & 46.00 & 0.91 \\
\midrule
child4-1-greedy & 50.56 & 68.46 & 12.20 & 43.74 & 0.79 \\
child4-2-greedy & 51.28 & 68.46 & 19.51 & 46.42 & 0.95 \\
\midrule
child5-1-greedy & 51.36 & 68.69 & 20.73 & 46.93 & 0.99 \\
child5-2-greedy & 50.49 & 73.24 & 9.76 & 44.50 & 0.78 \\
\midrule
child6-1-greedy & 50.50 & 73.24 & 9.15 & 44.30 & 0.78 \\
child6-2-greedy & 51.42 & 69.14 & 20.12 & 46.89 & 0.99 \\
\midrule
child7-1-greedy & 51.36 & 68.82 & 20.34 & 46.84 & 0.99 \\
child7-2-greedy & 51.42 & 68.74 & 20.81 & 46.99 & 0.99 \\
child7-3-greedy & 51.44 & 69.15 & 20.44 & 47.01 & 0.99 \\
\bottomrule
\end{tabular}
\caption{Evaluation Results of Additional Experiments of Mistral-7B.}
\label{tab:full1-alt}
\end{table*}

\begin{table*}[ht]
\centering
\begin{tabular}{l|ccc|c|c}
\toprule
\textbf{Model} & \textbf{TruthfulQA} & \textbf{Winogrande} & \textbf{GSM8K} & \textbf{Avg.} & \textbf{Model Kinship} \\ 
\midrule
winogrande & 42.0 & 77.9 & 6.4 & 42.1 & / \\
GSM8K & 39.0 & 73.4 & 38.0 & 50.1 & / \\
TruthfulQA & 56.7 & 68.9 & 9.5 & 45.0 & / \\
\midrule
child1-1-greedy & 40.2 & 79.3 & 34.2 & 51.2 & 0.03 \\
child1-2-greedy & 46.7 & 74.4 & 34.2 & 51.7 & 0.01 \\
child1-3-greedy & 46.1 & 77.1 & 1.9 & 41.7 & 0.01 \\
\midrule
child-2-1-greedy & 44.5 & 78.5 & 36.8 & \textbf{\textcolor{red}{53.2}} & 0.19 \\
child-2-2-greedy & 43.7 & 74.0 & 40.4 & 52.7 & 0.45 \\
child-2-3-greedy & 38.9 & 77.5 & 37.1 & 51.1 & 0.39 \\
\textbf{child-2-exp} & 43.3 & 81.2 & 28.5 & 51.0 & 0.01 \\
\midrule
child-3-1-greedy & 44.2 & 77.1 & 37.3 & 52.8 & 0.88 \\
child-3-2-greedy & 45.4 & 77.5 & 34.5 & 52.4 & 0.79 \\
child-3-3-greedy & 45.0 & 73.8 & 36.6 & 51.8 & 0.89 \\
\textbf{child-3-exp} & 45.1 & 78.6 & 30.3 & 51.3 & 0.58 \\
\midrule
child-4-1-greedy & 44.4 & 78.5 & 36.8 & 53.2 & 0.95 \\
child-4-2-greedy & 44.1 & 75.5 & 40.0 & 53.1 & 0.97 \\
\textbf{child-4-exp} & 43.3 & 80.9 & 32.6 & 52.2 & 0.81 \\
\midrule
child-5-1-greedy & 44.2 & 77.1 & 37.2 & 52.8 & 0.97 \\
child-5-2-greedy & 44.3 & 77.4 & 36.7 & 52.8 & 0.91 \\
child-5-3-greedy & 44.3 & 78.3 & 36.8 & 53.1 & 0.98 \\
\textbf{child-5-exp} & 44.5 & 78.1 & 32.0 & 51.5 & 0.64 \\
\midrule
child-6-1-greedy & 44.5 & 78.5 & 36.8 & 53.2 & 0.99 \\
child-6-2-greedy & 44.4 & 78.3 & 36.8 & 53.2 & 0.99 \\
child-6-3-greedy & 44.3 & 78.3 & 36.8 & 53.1 & 0.99 \\
\textbf{child-6-exp} & 44.3 & 80.4 & 35.3 & \textbf{\textcolor{darkgreen}{53.4}} & 0.80 \\
\bottomrule
\end{tabular}
\caption{Evaluation Results of addtional experiments of Llama-2.}
\label{tab:full2}
\end{table*}

\clearpage

\section{Algorithm Details for the Main Experiment}
\label{sec:appendix-greedy}
In this section, we present the detailed algorithms employed in our main experiment, along with an ablation study to validate our baseline method, Top $k$ Greedy Merging.

\subsection{Algorithms}

The \textbf{Top-$k$ Greedy Merging} algorithm aims to iteratively construct improved models through pairwise merging, guided by performance evaluation and, in the extended version, model kinship. 
It begins with a set of $n$ foundation models $M = \{m_1, m_2, \dots, m_n\}$. 
In the first step, all possible pairs of models are merged to form the first generation $G_1$ of merged models. 
These new models are added back into the candidate set $M$.

The algorithm then evaluates all models in $M$ using a task-specific evaluation function $f$ and selects the top $k$ performing models to form the working set $S$. 
It maintains an iterative loop that continues as long as the top-$k$ set $S$ changes between iterations, ensuring exploration continues only while performance improves.
Within each iteration, all model pairs from $S$ are merged to produce the next generation of models $G_i$. These new models are added into $M$, and their performance is evaluated using $f$ to update $S$.

In the variant \textcolor{blue}{with model kinship}, additional steps introduce a model exploration mechanism. 
This kinship-guided exploration step is designed to escape local optima by encouraging diversity in the merging path, potentially leading to models with better generalization or complementary capabilities. 
The algorithm terminates when the top-$k$ set $S$ stabilizes, indicating no further performance gains are observed.
Each model is named according to its generation and creation order to track its origin during analysis.

\subsection{Ablation Study of Greedy Strategy}
\label{sec:appendix-random}
The ablation study on the Greedy Strategy is conducted using the Mistral-7B architecture, following the same experimental settings outlined in the main experiments. 
For comparison, we employ the \textbf{random-merge strategy}, where models in each generation are merged with randomly selected models (excluding themselves) from the repository, as illustrated in Algorithm \ref{alg:random_merge}.

The following table presents the evaluation results. Each column represents:

\begin{itemize}
    \item \textbf{Model:} The name of each model. Note that the first three entries are fine-tuned foundation models used in our experiments.
    \item \textbf{TruthfulQA\_mc2, Winogrande, GSM8K:} The benchmark results for each dataset, indicating the model's task-specific capabilities.
    \item \textbf{Average:} The average score across all benchmarks, reflecting the model's overall generalization performance.
    \item \textbf{Model Kinship:} The kinship score (Here, we use cosine similarity to measure model kinship) of the parent models involved in the merge, indicating their relatedness.
    \item \textbf{Parent-1 and Parent-2:} The names of the parent models used in the merging process.
\end{itemize}

In the \textbf{random-merge strategy}, the average performance in each generation fluctuates. 
The highest average performance achieved is 68.55, slightly lower than the 68.72 observed in the \textbf{greedy experiment}. 
While the random-merge strategy avoids convergence to local optima, it demonstrates an unstable improvement process, which can lead to unpredictable results.

\clearpage

\begin{table*}[ht]
\centering
\small
\begin{tabular}{l|cccc|c}
\toprule
\textbf{Model} & \textbf{TruthfulQA\_mc2} & \textbf{Winogrande} & \textbf{GSM8K} & \textbf{Average} & \textbf{Model Kinship} \\
\midrule
\text{MetaMath-mistral-7B} & 44.89 & 75.77 & 70.51 & 63.72 & / \\
\text{Mistral-7B-Instruct-v0.2} & 68.26 & 77.19 & 40.03 & 61.82 & / \\
\text{Open-chat-3.5-1210} & 52.15 & 80.74 & 65.96 & 66.28 & / \\
\midrule
\text{child1-1} & 52.51 & 76.16 & 57.85 & 62.17 & 0.01 \\
\text{child1-2} & 58.04 & 76.32 & 57.72 & 64.02 & 0.01 \\
\text{child1-3} & 48.96 & 78.69 & 72.86 & 66.84 & 0.03 \\
\midrule
\text{child2-1} & 44.68 & 74.00 & 50.80 & 56.40 & 0.29 \\
\text{child2-2} & 49.78 & 78.93 & 55.72 & 61.47 & 0.41 \\
\text{child2-3} & 61.01 & 79.56 & 63.76 & 68.11 & 0.01 \\
\midrule
\text{child3-1} & 51.52 & 78.23 & 56.71 & 62.15 & 0.84 \\
\text{child3-2} & 43.52 & 75.22 & 47.43 & 55.39 & 0.59 \\
\text{child3-3} & 54.32 & 78.53 & 72.81 & \textbf{68.55} & 0.28 \\
\midrule
\text{child4-1} & 55.32 & 78.41 & 56.23 & 63.32 & 0.54 \\
\text{child4-2} & 50.53 & 78.42 & 57.65 & 62.20 & 0.86 \\
\text{child4-3} & 53.45 & 79.31 & 72.65 & 68.47 & 0.67 \\
\bottomrule
\end{tabular}
\label{tab:random}
\caption{Evaluation results using the random-merge strategy.}
\end{table*}

\begin{algorithm*}
\caption{Top $k$ Greedy Merging \textcolor{blue}{with Model Kinship}.}
\label{alg:greedy_merge}
\begin{algorithmic}[1]
\REQUIRE A set $M$ of $n$ foundation models $\{m_1, m_2, \dots, m_n\}$, Evaluation function $f$, Similarity metric function $sim(\cdot, \cdot)$ for model kinship.
\STATE Generate the first generation of merged models $G_{1}$ by merging each pair in set $M$, and set generation $i$ = 1.
\STATE Combine the set $G_{1}$ into set $M$.
\STATE Evaluate each model $m$ in set $M$, and select the top $k$ models. Denote this set as $S = \{m_1, m_2, \dots, m_k\}$.
\STATE Initialize a variable $S_{\text{prev}} = \emptyset$ to store the top $k$ models from the previous iteration.
\WHILE{$S \neq S_{\text{prev}}$}
    \STATE i++
    \STATE Set $S_{\text{prev}} = S$.
    \STATE Select each model pair from $S$. Denote this set as $P = \{p_1, p_2, \dots, p_j\}$.
    \STATE Merge every selected pair in set $P$ as merged model set $G_i = \{m_1, m_2, \dots, m_j\}$ for generation $i$, and add each merged model into set $M$.
    \STATE \textcolor{blue}{Identify the current best model $m_{best} \in S$}.
    \STATE \textcolor{blue}{Identify the model $m_f \in S$ with the lowest model kinship to $m_{best}$ from the $G_{i-1}$ according to the similarity metric $sim(\cdot, \cdot)$}.
    \STATE \textcolor{blue}{Merge $m_f$ with $m_{best}$ to generate a new model $m_{\text{exp}}$, and add $m_{\text{exp}}$ into set $G_i$ and set $M$}.
    \STATE Evaluate each new model $m \in G_i$ using $f$ and update $S$.
    \STATE \textcolor{blue}{Evaluate $m_{\text{exp}}$ using $f$ and update $S$}.
\ENDWHILE
\end{algorithmic}
\begin{flushleft}
\textbf{Note:} The \textcolor{blue}{blue-highlighted} steps are only executed in modified experiments incorporating model kinship-based exploration. To distinguish between different models in the subsequent experiments, each model generated in a given generation is named as \textbf{model-generation-id}.
\end{flushleft}
\end{algorithm*}

\begin{algorithm*}
\caption{Random Merge Algorithm.}
\label{alg:random_merge}
\begin{algorithmic}[1]
\REQUIRE A set $M$ of $n$ foundation models $\{m_1, m_2, \dots, m_n\}$, Evaluation function $f$.
\STATE Generate the first generation of merged models $G_{1}$ by randomly merging pairs in set $M$, and set generation $i$ = 1.
\STATE Combine the set $G_{1}$ into set $M$.
\STATE Evaluate each model $m$ in set $M$.
\STATE Initialize a variable $S_{\text{prev}} = \emptyset$ to store the top $k$ models from the previous iteration.
\WHILE{$S \neq S_{\text{prev}}$}
    \STATE i++
    \STATE Set $S_{\text{prev}} = S$.
    \STATE Randomly select pairs of models from $M$. Denote this set as $P = \{p_1, p_2, \dots, p_j\}$.
    \STATE Merge each selected pair in set $P$ to form the merged model set $G_i = \{m_1, m_2, \dots, m_j\}$ for generation $i$, and add each merged model into set $M$.
    \STATE Evaluate each new model $m \in G_i$ using $f$ and update $S$.
\ENDWHILE
\end{algorithmic}
\end{algorithm*}

\clearpage
\section{Additional Analysis for Community Model Evolution}

\subsection{Analysis of Model Kinship Change across Merging Stages}
\label{sec:appendix-stage}
To determine whether the discovery of increasing model kinship in model evolution paths can be generalized to the entire model evolution process, we perform an analysis of the merging stages.
Given the community's predominant use of the performance-prior strategy, we calculate model kinship among models with similar performance, simulating the selection of models at each stage. 
For this analysis, we randomly select 5 models from each merging stage, as delineated by boundaries identified in prior analysis - Saturation Stage ($\geq0.75$), Improving Stage ($<$0.75 and $\geq$0.73), and Initial Merges (fine-tuned models) to form three foundation model groups, representing potential merges at different stages of model evolution.

\subsection{Details of Model Group Selection}
\label{sec:appendix-group}

This section presents the exact models included in each model group, as shown in Table \ref{tab:exp3}. 
The selection process is conducted across three distinct groups: \textbf{(1)} the top 5 models on the leaderboard, with a performance difference of 0.2, \textbf{(2)} 5 models with performance scores around 73 and a performance difference of 0.2, and \textbf{(3)} 5 fine-tuned models. 
It is important to note that the fine-tuned models are not selected based on performance, and may exhibit significant differences in results.

\begin{table}[!htbp]
\small
\begin{center}
\begin{tabular}{c|c}
    \toprule
    \textbf{Group} & \textbf{Models}\\
    \midrule
    \multirow{5}{*}{Top Model Group}& YamshadowExperiment28-7B\\
    & Yamshadow-7B\\
    & Experiment25-7B\\
    & StrangeMerges\_24-7B-slerp\\
    & MonaTrix-v6\\
    \hline
    \multirow{5}{*}{Mid Stage Model Group}& Daredevil-7B\\
    & CatMarcoro14-7B\\
    & Mayo\\
    & Calmesmol-7B-slerp \\
    & StrangeMerges\_4-7B-slerp\\
    \hline
    \multirow{5}{*}{Fine-tuned Model Group}& Zephyr-beta\\
    & MetaMath-Mistral-7B\\
    & Mistral-7B-Instruct-v0.2\\
    & openchat-3.5-1210\\
    & WizardLM-2\\
    \bottomrule
\end{tabular}
\caption{Model Group in Kinship Matrix Analysis.}
\label{tab:exp3}
\end{center}
\end{table}

\begin{figure}[ht]
    \centering
    \includegraphics[width=\linewidth]{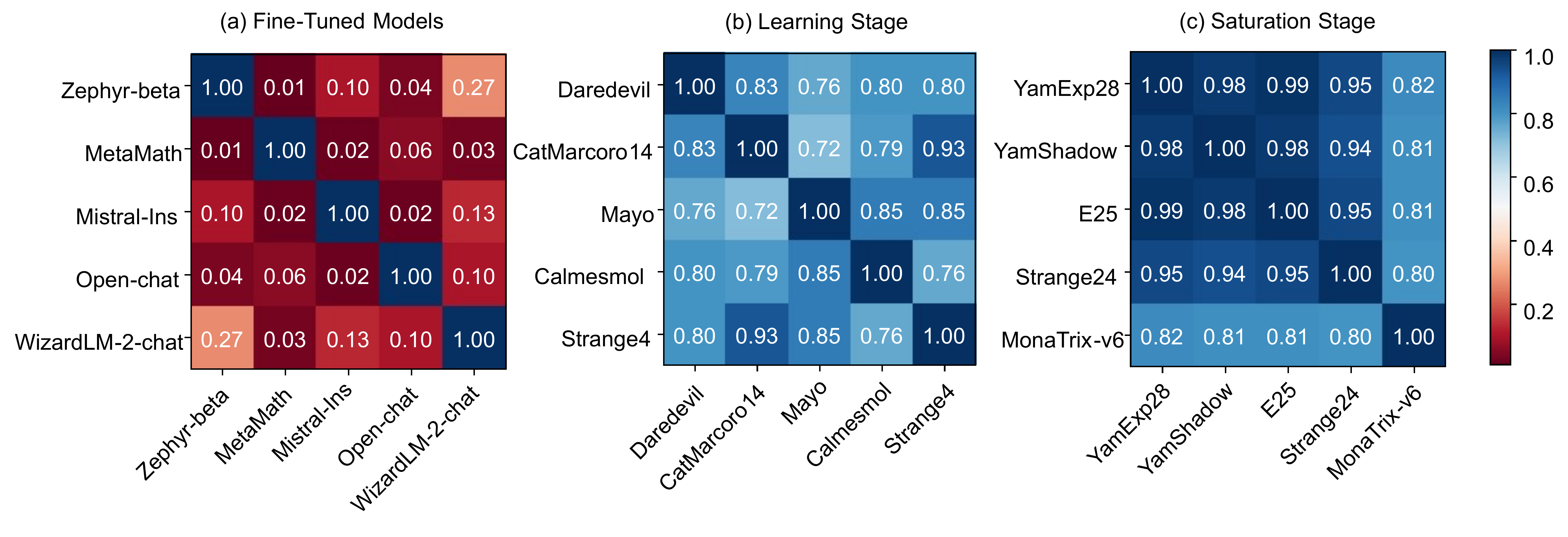}
    \caption{\textbf{
    The Model Kinship Matrices for the three model groups}. Each element represents the model kinship value between the corresponding models. 
    In Group B and C, the merged models are arranged by average task performance, ordered from \textbf{high to low} (left to right).}
    \label{fig:matrix}
\end{figure}

Figure \ref{fig:matrix} illustrates the model kinship between models within each group. 
We observe that model kinship increases with the average task performance across models that follow different evolution paths. 
Additionally, during the saturation stage, all potential merges display a strong affinity, with model kinship values nearing 1.

\clearpage
\section{Analysis between Task Relatedness and Model Kinship}
\label{sec:appendix-task}
In the formulation of model kinship, we use the placeholder \( \sim(\cdot, \cdot) \) as a similarity metric function to explore options that can effectively capture task-related differences. 
One such metric is cosine similarity, derived from the analysis in the task vector, which has been validated as effective for representing differences in single-task models through the cosine similarity of delta parameters (task vectors). In addition to cosine similarity, we also investigate the Pearson correlation coefficient and Euclidean distance.

However, we have not thoroughly evaluated the applicability of these metrics in the context of model evolution, particularly for merged models with multi-task capabilities. 
To address this, we examine the relationship between the similarity metrics and task information in subsequent sections.

Our analysis focuses on the LLaMA-2 architecture, as we can find the necessary open-source fine-tuned checkpoints on various datasets. 
To measure differences between models, we currently use a preliminary evaluation method: the \textbf{Average Task Performance Difference} (ATPD), which aims to represent task capability differences based on evaluation performance.

The Average Task Performance Difference (ATPD) between two models, \( M_1 \) and \( M_2 \), is calculated by averaging the absolute differences in performance across all tasks. Let \( T \) denote the set of tasks, and \( P_i^{(j)} \) represent the performance of model \( M_j \) on task \( i \).  
Then, the ATPD is defined as:

\[
\text{ATPD}(M_1, M_2) = \frac{1}{|T|} \sum_{i \in T} \left| P_i^{(1)} - P_i^{(2)} \right|
\]

\begin{itemize}
    \item \( |T| \): the total number of tasks.
    \item \( P_i^{(1)} \) and \( P_i^{(2)} \): performances of models \( M_1 \) and \( M_2 \) on task \( i \).
    \item \( \left| P_i^{(1)} - P_i^{(2)} \right| \): absolute difference in performance for task \( i \).
\end{itemize}

\begin{table}[h!]
\centering
\small
\begin{tabular}{l|c|c|c}
\hline
\textbf{Method} & \textbf{Corr(cs)} & \textbf{Corr(pcc)} & \textbf{Corr(ed)} \\ \hline
Value                   & -0.77                    & -0.74                     & 0.80                     \\ \hline
\end{tabular}
\caption{Correlation values between ATPD and model kinship.}
\label{tab:correlation}
\end{table}

For this study, we utilize models from additional \textbf{LLaMA-2} experiments (Appendix.\ref{sec:appendix-full}). These models are merged from three fine-tuned models, allowing us to control the generated models to focus solely on the corresponding task capabilities. The following table presents the results, with Winogrande, TruthfulQA, and GSM8K representing the performance differences across each task.

The results in Table.\ref{tab:correlation} demonstrate strong correlations: Cosine Similarity (-0.77) and Pearson Correlation Coefficient (-0.74) exhibit negative correlations, while Euclidean Distance (0.80) shows a positive correlation. This supports that model kinship is related to task differences. As mentioned in the limitations, the current metrics are viable but not optimal. Combining them with task information studies could further enhance the value of our work.

\subsection{Additional Results: Analysis of Model Kinship and Average Task Performance}
\begin{figure*}[ht]
    \centering
    \fontsize{8}{10}\selectfont
    \includegraphics[width=\linewidth]{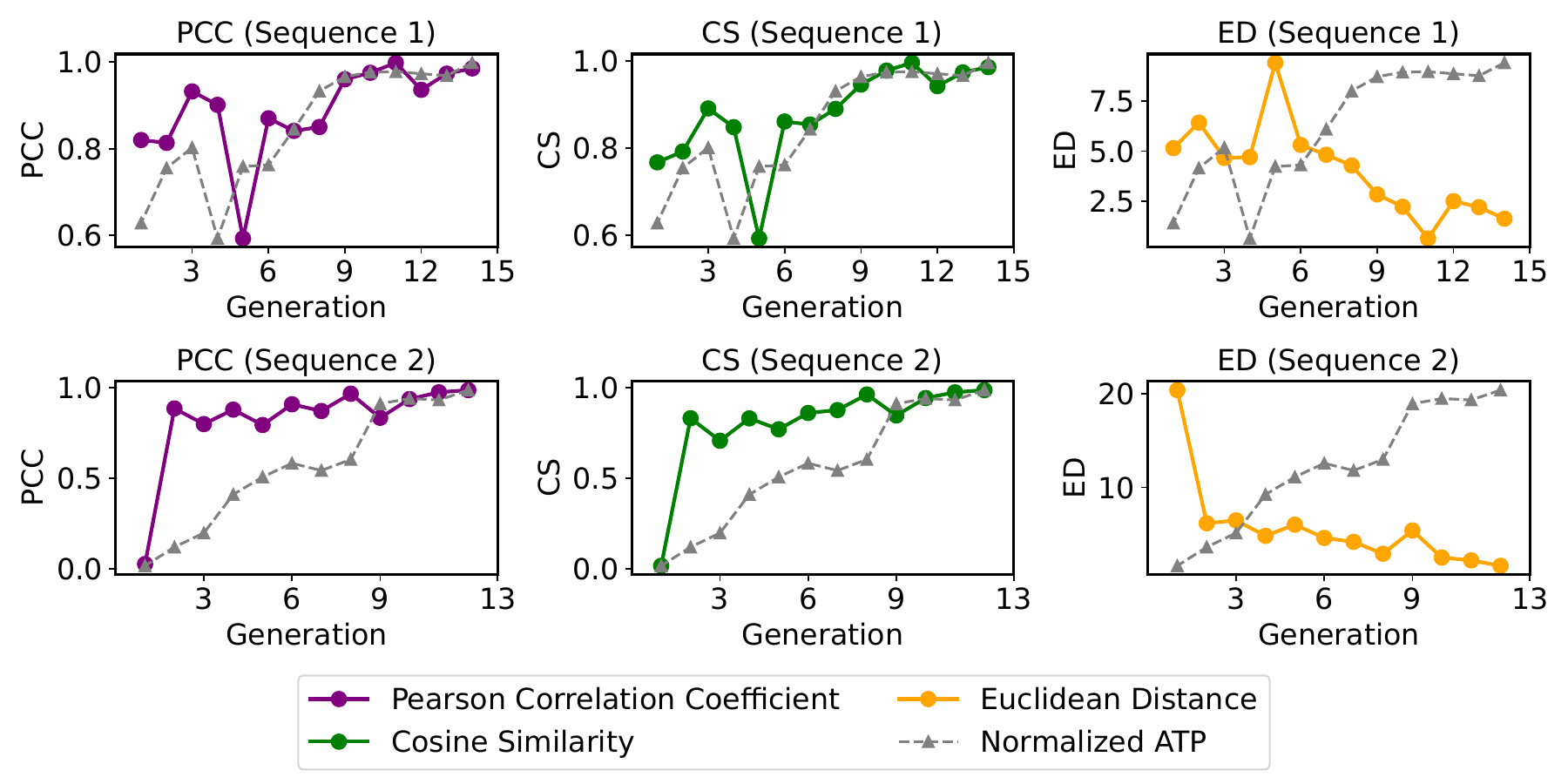}
    \caption{Illustration of comparison between the correlation of Pearson Correlation Coefficient (PCC), Cosine Similarity (CS), and Euclidean Distance (ED) with average task performance (Normalized to the same value scale).}
    \label{fig:mk_atp}
\end{figure*}

This section provides supplementary analysis on the relationship between model kinship and average task performance. Figure \ref{fig:mk_atp} illustrates a comparison between average task performance and model kinship using two additional metrics not included in the main paper. From an intuitive observation, model kinship based on the three metrics exhibits a similar correlation with average task performance.
\clearpage

\begin{table*}[ht]
  \centering
  \tiny
  \begin{tabular}{l l | c c c c | c c c}
    \toprule
    \textbf{Model 1} & \textbf{Model 2} & \textbf{Winogrande} & \textbf{TruthfulQA} & \textbf{GSM8K} & \textbf{ATPD} & \textbf{Kinship(cs)} & \textbf{Kinship(pcc)} & \textbf{Kinship(ed)} \\
    \midrule
    \text{child-4-1-greedy} & \text{child-5-3-greedy} & 0.10 & 0.00 & 0.20 & 0.10 & 0.99 & 0.99 & 2.17 \\
    \text{child-2-1-greedy} & \text{child-4-1-greedy} & 0.20 & 0.10 & 0.00 & 0.10 & 0.98 & 0.99 & 4.22 \\
    \text{child-2-1-greedy} & \text{child-5-3-greedy} & 0.10 & 0.10 & 0.20 & 0.13 & 0.99 & 0.99 & 2.19 \\
    \text{child-4-exp} & \text{child-2-1-greedy} & 1.10 & 0.90 & 0.10 & 0.70 & 0.80 & 0.75 & 25.53 \\
    \text{child-2-1-greedy} & \text{child-3-1-greedy} & 0.20 & 1.30 & 0.70 & 0.73 & 0.95 & 0.98 & 6.74 \\
    \text{child-4-1-greedy} & \text{child-6-exp} & 0.10 & 1.90 & 1.40 & 1.13 & 0.74 & 0.71 & 25.54 \\
    \text{child-4-1-greedy} & \text{child-4-2-greedy} & 0.30 & 3.00 & 3.20 & 2.17 & 0.97 & 0.98 & 6.57 \\
    \text{child-2-2-greedy} & \text{child-3-1-greedy} & 0.50 & 3.10 & 3.10 & 2.23 & 0.97 & 0.98 & 6.57 \\
    \text{child-2-1-greedy} & \text{child-4-2-greedy} & 0.50 & 3.10 & 3.20 & 2.27 & 0.91 & 0.96 & 9.29 \\
    \text{child-3-exp} & \text{child-2-1-greedy} & 0.70 & 0.20 & 6.30 & 2.40 & 0.64 & 0.52 & 35.52 \\
    \text{child-4-exp} & \text{child-2-1-greedy} & 1.10 & 2.50 & 4.00 & 2.53 & 0.78 & 0.75 & 25.53 \\
    \text{child-2-1-greedy} & \text{child1-2-greedy} & 2.30 & 4.00 & 2.40 & 2.90 & 0.79 & 0.89 & 15.75 \\
    \text{child-2-1-greedy} & \text{child-2-2-greedy} & 0.70 & 4.40 & 3.80 & 2.97 & 0.88 & 0.95 & 12.43 \\
    \text{child-2-2-greedy} & \text{child1-2-greedy} & 3.00 & 0.40 & 6.20 & 3.20 & 0.89 & 0.92 & 11.68 \\
    \text{child1-1-greedy} & \text{GSM8K} & 1.20 & 5.90 & 3.80 & 3.63 & 0.39 & 0.46 & 36.39 \\
    \text{child1-1-greedy} & \text{child1-2-greedy} & 6.50 & 4.90 & 0.00 & 3.80 & 0.19 & 0.16 & 38.07 \\
    \text{child-2-exp} & \text{child-2-1-greedy} & 1.10 & 2.80 & 8.10 & 4.00 & 0.58 & 0.77 & 28.33 \\
    \text{child1-2-greedy} & \text{GSM8K} & 7.70 & 1.00 & 3.80 & 4.17 & 0.45 & 0.38 & 26.32 \\
    \text{child-2-1-greedy} & \text{child1-3-greedy} & 7.80 & 3.10 & 2.90 & 4.60 & 0.58 & 0.51 & 45.24 \\
    \text{child-3-1-greedy} & \text{child-2-exp} & 0.90 & 4.10 & 8.80 & 4.60 & 0.58 & 0.63 & 32.45 \\
    \text{winogrande} & \text{TruthfulQA} & 14.70 & 9.00 & 3.10 & 8.93 & 0.01 & 0.01 & 74.49 \\
    \text{child1-2-greedy} & \text{child1-3-greedy} & 0.60 & 2.70 & 32.30 & 11.87 & 0.64 & 0.52 & 46.06 \\
    \text{child1-2-greedy} & \text{winogrande} & 4.70 & 3.50 & 27.80 & 12.00 & 0.01 & 0.02 & 55.89 \\
    \text{winogrande} & \text{GSM8K} & 3.00 & 4.50 & 31.60 & 13.03 & 0.03 & 0.11 & 54.01 \\
    \text{child1-1-greedy} & \text{child1-3-greedy} & 5.90 & 2.20 & 32.30 & 13.47 & 0.52 & 0.64 & 44.16 \\
    \text{GSM8K} & \text{TruthfulQA} & 17.70 & 4.50 & 28.50 & 16.90 & 0.01 & 0.01 & 61.56 \\
    \bottomrule
  \end{tabular}
  \label{tab:main_sum}
\caption{Summary of Model Merging Results.}
\end{table*}

\label{sec:appendix-c}

\clearpage
\section{Optimization Assumption of Model Evolution}
\label{sec:appendix-b}
\begin{figure}[H]
\includegraphics[width=0.8\linewidth]{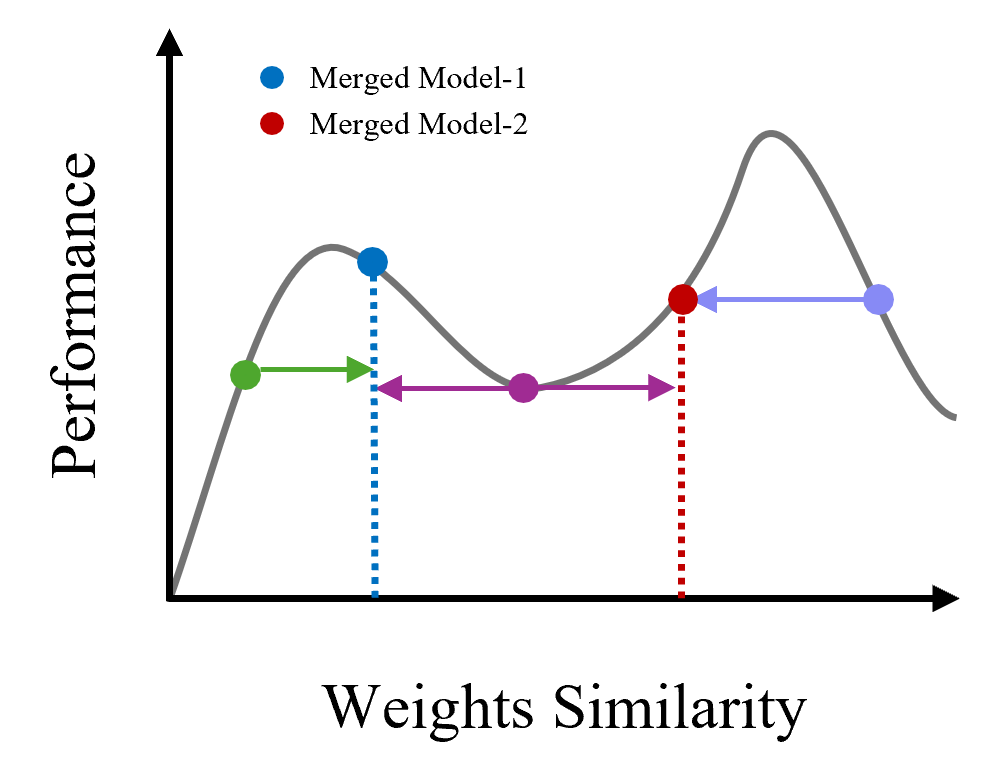} 
\caption{An intuitive illustration of \textbf{how model evolution can fall into local optima} due to a performance-prior strategy. It shows that Merged Model 2 may be overlooked, despite its potential for better multi-task performance.}
\label{fig:add}
\end{figure}
Our findings using new strategy offer a new perspective on model evolution through multiple merging. If the merging process can be improved using a common optimization strategy, it raises the question of \textit{whether the underlying mechanism mirrors this optimization problem}. 
Thus, we hypothesize the following:

\intuition{\textbf{Hypothesis:} The evolution process may be simplified to a binary search process for most weight-averaging-based model merging methods.}

\begin{figure*}[ht]
    \centering
    \includegraphics[width=\linewidth]{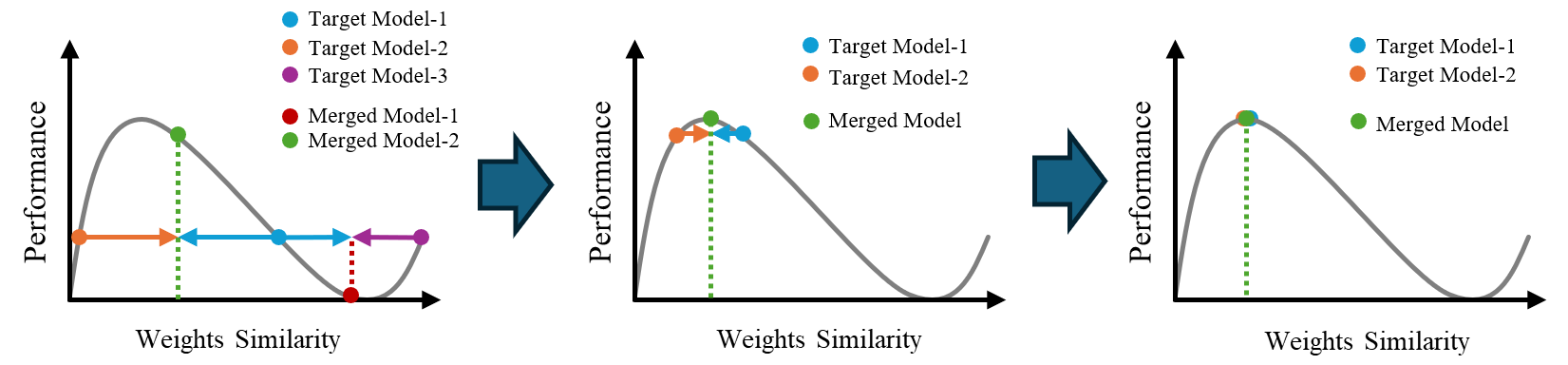}
    \caption{An intuitive illustration of \textbf{the optimization process assumption} in model evolution, where models progressively converge towards the optimal model.}
    \label{fig:ini}
\end{figure*}

Figure \ref{fig:ini} illustrates the ideal scenario in our assumption where multiple merges produce a highly generalized model. For the generalization task $t$, the y-axis represents the model performance for task $t$ and the x-axis represents the model's weight space. In early merging stages, models fine-tuned with different tasks exhibit significant weight space dissimilarity. The merging process averages these weight spaces, and the experiment conductor selects the better-merged models while discarding the inferior ones. In stage 2, the search area narrows and the improvements become stable, eventually leading to an optimized state in stage 3 when ``saturation stage'' occurs.

In this context, Model Kinship serves as a metric to quantify the weight space distance between two models, with a higher model kinship indicating a lower weight space distance. Following this assumption, our findings of the optimization problem in model evolution can be elucidated in Figure \ref{fig:add}.

However, we currently lack sufficient evidence to validate this hypothesis. 
Future work is needed to explore this further.

\clearpage

\begin{figure*}[ht]
    \centering
    \includegraphics[width=\linewidth]{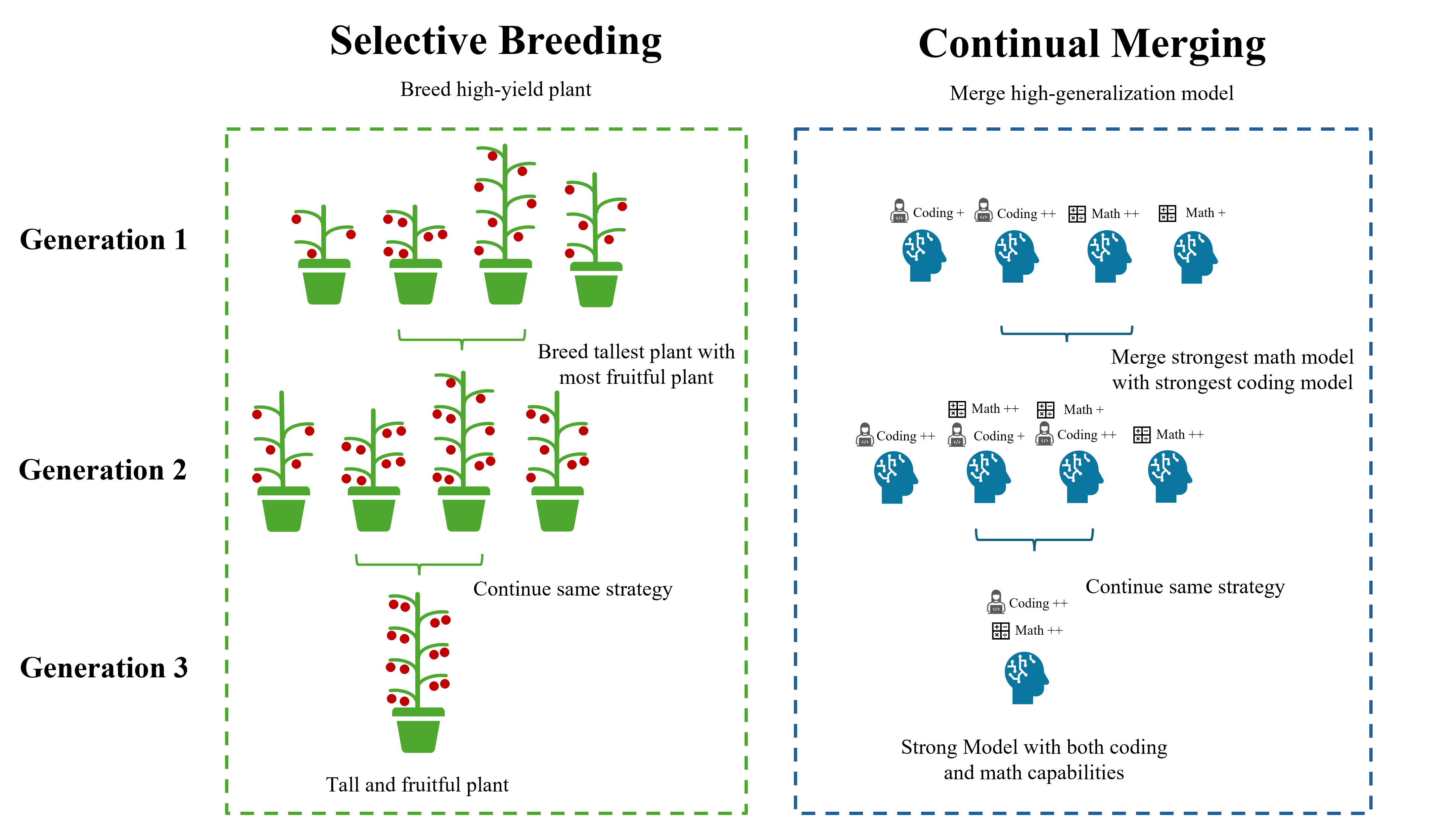} 
    \caption{An intuitive \textbf{comparison between selective
breeding and iterative model merging}. The \textbf{left} process
demonstrates breeding a tall and frutful plant by selecting parents with the desired traits in an biological scenario. The \textbf{right} process shows developing a model with capabilities of coding and math through model evolution.}
    \label{fig:other2}
\end{figure*}

\clearpage
\section{Referenced Concepts in Evolutionary biology}
\label{biology}

In this section, we detail the conceptual parallels between biological processes and model merging, highlighting our motivation for employing model kinship.

\subsection{Iterative Merging vs. Artificial Selection}
We draw inspiration for model evolution from biological evolution, specifically focusing on the correlation between biological evolution through artificial selection and model evolution via model merging. Artificial selection involves retaining desirable traits by manually selecting breeding pairs in each generation, typically those exhibiting the most significant features. Similarly, model evolution, as explored in this paper through Iterative Model Merging, adopts a comparable approach: users preserve desired task capabilities by strategically selecting merging pairs. Through iterative merging, they can develop a model that is proficient in all tasks in a given task set. To illustrate this comparison more effectively, Figure \ref{fig:other2} shows an example of combining two features/task capabilities in evolution.

\subsection{Inbreeding Depression vs. Saturation Stage}

As highlighted in the main paper, one of our key findings is that the late stage of model evolution often enters a saturation stage, during which models exhibit minimal differences from one another. This phenomenon parallels "inbreeding depression" in artificial selection, where breeding closely related individuals reduces genetic diversity and fitness. Although genetic inheritance and model weights operate differently, merging closely related models leads to new models with minimal variation, thereby reducing the effectiveness of merging, particularly in weight averaging. To address this issue, we propose quantifying the differences between models, a concept we term model kinship, to guide the merging process and mitigate the challenges associated with the saturation stage in model evolution.

\section{References to Open Models}
\label{sec:appendix-ref}
See Table~\ref{tab:ref}.

\onecolumn
\small
\begin{longtable}{l|l}
\toprule
\textbf{Model Name} & \textbf{HuggingFace Reference} \\ 
\midrule
Multi\_verse\_model-7B & \url{https://huggingface.co/MTSAIR/multi_verse_model} \\
Experiment26-7B & \url{https://huggingface.co/yam-peleg/Experiment26-7B} \\
M7-7b & \url{https://huggingface.co/liminerity/M7-7b} \\
StrangeMerges\_32-7B-slerp & \url{https://huggingface.co/Gille/StrangeMerges_32-7B-slerp} \\
Ognoexperiment27 & \url{https://huggingface.co/automerger/OgnoExperiment27-7B} \\
YamShadow-7B & \url{https://huggingface.co/automerger/YamShadow-7B} \\
Experiment28 & \url{https://huggingface.co/yam-peleg/Experiment28-7B} \\
shadow-clown-7B-slerp & \url{https://huggingface.co/CorticalStack/shadow-clown-7B-slerp} \\
yam-jom-7B & \url{https://huggingface.co/mayacinka/yam-jom-7B} \\
StrangeMerges\_21-7B-slerp & \url{https://huggingface.co/Gille/StrangeMerges_21-7B-slerp} \\
StrangeMerges\_31-7B-slerp & \url{https://huggingface.co/Gille/StrangeMerges_31-7B-slerp} \\
NeuralBeagle14-7B & \url{https://huggingface.co/mlabonne/NeuralBeagle14-7B} \\
Turdus & \url{https://huggingface.co/udkai/Turdus} \\
DareBeagle-7B & \url{https://huggingface.co/shadowml/DareBeagle-7B} \\
TurdusBeagle-7B & \url{https://huggingface.co/leveldevai/TurdusBeagle-7B} \\
FernandoGPT-v1 & \url{https://huggingface.co/samir-fama/FernandoGPT-v1} \\
StrangeMerges\_10-7B-slerp & \url{https://huggingface.co/Gille/StrangeMerges_10-7B-slerp} \\
TurdusDareBeagle-7B & \url{https://huggingface.co/leveldevai/TurdusDareBeagle-7B} \\
MarcMistral-7B & \url{https://huggingface.co/flemmingmiguel/MarcMistral-7B} \\
StrangeMerges\_20-7B-slerp & \url{https://huggingface.co/Gille/StrangeMerges_20-7B-slerp} \\
NeuTrixOmniBe-7B-model-remix & \url{https://huggingface.co/Kukedlc/NeuTrixOmniBe-7B-model-remix} \\
StrangeMerges\_11-7B-slerp & \url{https://huggingface.co/Gille/StrangeMerges_11-7B-slerp} \\
MBX-7B-v3 & \url{https://huggingface.co/flemmingmiguel/MBX-7B-v3} \\
Marcoroni-7B-v3 & \url{https://huggingface.co/AIDC-ai-business/Marcoroni-7B-v3} \\
Mistral-7B-Merge-14-v0.1 & \url{https://huggingface.co/EmbeddedLLM/Mistral-7B-Merge-14-v0.1} \\
distilabeled-Marcoro14-7B-slerp & \url{https://huggingface.co/argilla/distilabeled-Marcoro14-7B-slerp} \\
UNA-TheBeagle-7b-v1 & \url{https://huggingface.co/fblgit/UNA-TheBeagle-7b-v1} \\
CatPPT-base & \url{https://huggingface.co/rishiraj/CatPPT-base} \\
CatMacaroni-Slerp & \url{https://huggingface.co/cookinai/CatMacaroni-Slerp} \\
LeoScorpius-7B & \url{https://huggingface.co/viethq188/LeoScorpius-7B} \\
NeuralDaredevil-7B & \url{https://huggingface.co/mlabonne/NeuralDaredevil-7B} \\
StrangeMerges\_9-7B-dare\_ties & \url{https://huggingface.co/Gille/StrangeMerges_9-7B-dare_ties} \\
mistral-ft-optimized-1218 & \url{https://huggingface.co/OpenPipe/mistral-ft-optimized-1218} \\
NeuralHermes-Mistral-2.5-7B & \url{https://huggingface.co/mlabonne/NeuralHermes-2.5-Mistral-7B} \\
neural-chat-7b-v3-2 & \url{https://huggingface.co/Intel/neural-chat-7b-v3-2} \\
OpenHermes-2.5-Mistral-7B & \url{https://huggingface.co/teknium/OpenHermes-2.5-Mistral-7B} \\
StrangeMerges\_30-7B-slerp & \url{https://huggingface.co/Gille/StrangeMerges_30-7B-slerp} \\
Experiment24 & \url{https://huggingface.co/yam-peleg/Experiment24-7B} \\
neural-chat-7b-v3-3 & \url{https://huggingface.co/Intel/neural-chat-7b-v3-3} \\
MultiverseEx26-7B-slerp & \url{https://huggingface.co/allknowingroger/MultiverseEx26-7B-slerp} \\
CalmExperiment-7B-slerp & \url{https://huggingface.co/allknowingroger/CalmExperiment-7B-slerp} \\
CapybaraMarcoroni-7B & \url{https://huggingface.co/AtAndDev/CapybaraMarcoroni-7B} \\
DistilHermes-2.5-Mistral-7B & \url{https://huggingface.co/eren23/DistilHermes-2.5-Mistral-7B} \\
Calme-7B-Instruct-v0.9 & \url{https://huggingface.co/MaziyarPanahi/Calme-7B-Instruct-v0.9} \\
StrangeMerges\_16-7B-slerp & \url{https://huggingface.co/Gille/StrangeMerges_16-7B-slerp} \\
coven\_7b\_128k\_orpo\_alpha & \url{https://huggingface.co/raidhon/coven_7b_128k_orpo_alpha} \\
Kunoichi-DPO-v2-7B & \url{https://huggingface.co/SanjiWatsuki/Kunoichi-DPO-v2-7B} \\
AlphaMonarch-7B & \url{https://huggingface.co/mlabonne/AlphaMonarch-7B} \\
StrangeMerges\_15-7B-slerp & \url{https://huggingface.co/Gille/StrangeMerges_15-7B-slerp} \\
Kunoichi-7B & \url{https://huggingface.co/SanjiWatsuki/Kunoichi-7B} \\
Mistral-T5-7B-v1 & \url{https://huggingface.co/ignos/Mistral-T5-7B-v1} \\
Marcoroni-neural-chat-7B-v2 & \url{https://huggingface.co/Toten5/Marcoroni-neural-chat-7B-v2} \\
Marcoro14-7B-slerp & \url{https://huggingface.co/Rupesh2/Marcoro14-7B-slerp} \\
MarcDareBeagle-7B & \url{https://huggingface.co/leveldevai/MarcDareBeagle-7B} \\
MarcBeagle-7B & \url{https://huggingface.co/leveldevai/MarcBeagle-7B} \\
MetaMath-Mistral-7B & \url{https://huggingface.co/meta-math/MetaMath-Mistral-7B} \\
openchat-3.5-1210 & \url{https://huggingface.co/openchat/openchat-3.5-1210} \\
Tulpar-7b-v2 & \url{https://huggingface.co/HyperbeeAI/Tulpar-7b-v2} \\
YugoGPT & \url{https://huggingface.co/gordicaleksa/YugoGPT} \\
\bottomrule
\caption{Model and Hugging Face Reference Links}\label{tab:ref} \\
\end{longtable}

\end{document}